# Blind Multiclass Ensemble Classification

Panagiotis A. Traganitis, *Student Member, IEEE,* Alba Pagès-Zamora, *Senior Member, IEEE,* and Georgios B. Giannakis, *Fellow, IEEE*

*Abstract*—The rising interest in pattern recognition and data analytics has spurred the development of innovative machine learning algorithms and tools. However, as each algorithm has its strengths and limitations, one is motivated to judiciously fuse multiple algorithms in order to find the "best" performing one, for a given dataset. Ensemble learning aims at such high-performance meta-algorithm, by combining the outputs from multiple algorithms. The present work introduces a blind scheme for learning from ensembles of classifiers, using a moment matching method that leverages joint tensor and matrix factorization. Blind refers to the combiner who has no knowledge of the ground-truth labels that each classifier has been trained on. A rigorous performance analysis is derived and the proposed scheme is evaluated on synthetic and real datasets.

## I. INTRODUCTION

THE massive amounts of generated and communicated data has introduced society and computing to a data "deluge." Along with the increase in the amount of data, multiple machine learning, signal processing and data mining algorithms have been developed. These algorithms are usually tailored for different datasets, and they often operate under different assumptions. As such, finding an algorithm that works "well" for a specific dataset can be prohibitively complex or impossible.

Ensemble learning refers to the task of designing a meta-learner, by combining the results provided by multiple different learners or annotators[1]; see Fig. 1. This meta-learner should generally be able to outperform the individual learners. In particular, ensemble classification refers to fusing the results provided by different classifiers. Each classifier observes data, decides one class, out of $K$ possible, each of these data belong to, and provides the meta-learner with those decisions. Such a setup emerges in diverse disciplines including medicine [1], biology [2], team decision making and economics [3], and has recently gained attention with the advent of crowdsourcing [4], as well as services such as Amazon's Mechanical Turk [5], CrowdFlower and Clickworker, to name a few. A related setup appears in distributed detection [6], [7], where sensors collect data, decide which one out of $K$ possible hypotheses is in effect, and transmit those decisions to a fusion center, that makes a final decision. A similar task is also known as the *CEO problem* or multiterminal source coding [8].

When training data are available, a meta-learner can learn how to combine the results from individual classifiers, based on these ground-truth labels [9]. One such approach is boosting [10], where multiple classifiers are combined according to their probability of error on the training set. In the boosting regime, each classifier is also using information from the rest. In many cases however, labeled data are not available to train the combining meta-classifier, and/or, the individual classifiers cannot be retrained, justifying the need for *unsupervised* (or *blind*) ensemble learning methods. One such paradigm is provided by crowdsourcing, where people are tasked with providing classification labels. Accordingly, in a distributed detection setup, the fusion center might not have access to the sensors, once they have been deployed.

The present work puts forth a novel scheme for *multiclass blind ensemble classification*, built upon simple concepts from probability and detection theory, as well as recent advances in tensor decompositions [11] and optimization theory, that enable assessing the reliability of multiple annotators and combining their answers. Under our proposed model, each annotator has a fixed probability of deciding that a datum belongs to class $k$, given that the true class of the datum is $k'$. Assuming that annotators make decisions independent of each other, the proposed method extracts these probabilities from the first-, second-, and third-order statistics of annotator responses. This becomes possible thanks to a joint PARAFAC decomposition, which has been employed in a related problem of identifying conditional probabilities to complete a joint probability functions from its projections [12]. The crux of our algorithm is a moment matching method, that leverages the aforementioned PARAFAC decomposition approach to obtain accurate estimates of annotator decision probabilities along with class priors. These estimates are then provided to the meta-detector to form the final estimate of data labels.

To assess the proposed scheme, extensive numerical tests, on synthetic as well as real data are presented, comparing the proposed approach to state-of-the-art binary and multiclass blind ensemble classification methods. In addition, a rigorous performance analysis is provided, which showcases the conditions under which our novel method works.

The rest of the paper is organized as follows. Section II states the problem, provides preliminaries for the proposed approach along with a brief description of the prior art in unsupervised ensemble classification. Section III introduces the proposed scheme for multiclass unsupervised ensemble classification, while Section IV analyses the performance of the proposed method. Section V presents numerical tests to compare our method with state-of-the-art ensemble classification algorithms. Finally, concluding remarks and future research directions are given in Section VI. Detailed deriva-

Panagiotis A. Traganitis and Georgios B. Giannakis are with the Dept. of Electrical and Computer Engineering and the Digital Technology Center, University of Minnesota, Minneapolis, MN 55455, USA.
Alba Pagès-Zamora is with the SPCOM Group, Universitat Politècnica de Catalunya BarcelonaTech, Spain.
Emails: {traga003@umn.edu, alba.pages@upc.edu, georgios@umn.edu}

[1]The terms learner, annotator, and classifier will be used interchangeably throughout this manuscript.

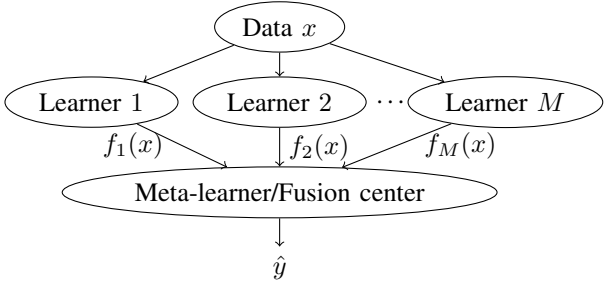

Fig. 1. Unsupervised ensemble classification setup, where the outputs of learners are combined in parallel.

tions are delegated to Appendix A, while proofs of theorems, propositions and lemmata are deferred to Appendix B.

**Notation:** Unless otherwise noted, lowercase bold letters, $\boldsymbol{x}$, denote vectors, uppercase bold letters, $\mathbf{X}$, represent matrices, and calligraphic uppercase letters, $\mathcal{X}$, stand for sets. The $(i,j)$th entry of matrix $\mathbf{X}$ is denoted by $[\mathbf{X}]_{ij}$; and its rank by $\text{rank}(\mathbf{X})$; $\mathbf{X}^\top$ denotes the tranpose of matrix $\mathbf{X}$; $\mathbb{R}^D$ stands for the $D$-dimensional real Euclidean space, $\mathbb{R}_+$ for the set of positive real numbers, $\mathbb{Z}_+$ for the set of positive integers, $\mathbb{E}[\cdot]$ for expectation, and $\|\cdot\|$ for the $\ell_2$-norm. Underlined capital letters $\underline{X}$ denote tensors, $\text{vec}(\cdot)$ denotes the vectorization operator, that stacks columns of a matrix into a longer column vector; the vector outer product is denoted by $\circ$, and, $\odot$ denotes the Khatri-Rao matrix product. For a 3-mode tensor $\underline{X}$, $\underline{X}(:,:,i), \underline{X}(:,i,:)$, and $\underline{X}(i,:,:)$ denote the $i$-th frontal, longitudinal and lateral slabs of $\underline{X}$, respectively, while $\underline{X}(i,j,l)$ denotes the $ijl$-th element of $\underline{X}$.

## II. PROBLEM STATEMENT AND PRELIMINARIES

Consider a dataset consisting of $N$ data (possibly vectors) $\{x_n\}_{n=1}^N$ each belonging to one of $K$ possible classes with corresponding labels $\{y_n\}_{n=1}^N$, e.g. $y_n = k$ if $x_n$ belongs to class $k$. The pairs $\{(x_n, y_n)\}$ are drawn independently from an unknown joint distribution D, and $X$ and $Y$ denote random variables such that $(X, Y) \sim$ D. Consider now $M$ annotators that observe $\{x_n\}_{n=1}^N$, and provide estimates of labels. Let $f_m(x_n) \in \{1, \ldots, K\}$ denote the label assigned to datum $x_n$ by the $m$-th annotator. All annotator responses are then collected at a centralized meta-learner or fusion center. The task of *unsupervised ensemble classification* is: Given only the annotator responses $\{f_m(x_n), m = 1, \ldots, M\}_{n=1}^N$, we wish to estimate the ground-truth labels of the data $\{y_n\}$; see Fig. 1.

Similar to unsupervised ensemble classification, crowd-sourced classification seeks to estimate ground-truth labels of the data $\{y_n\}$ from annotator responses $\{f_m(x_n)\}$, with the additional caveat that each annotator $m$ may choose to provide labels for only a subset $N_m < N$ of data.

### A. Prior work

Probably the simplest scheme for blind or unsupervised ensemble classification is majority voting, where the estimated label of a datum is the one that most annotators agree upon. Majority voting has been used in popular ensemble schemes such as bagging, and random forests [13]. While relatively easy to implement, majority voting presumes that all annotators are equally "reliable," which is rather unrealistic, both in crowdsourcing as well as in ensemble learning setups. Other blind ensemble methods aim to estimate the parameters that characterize the annotators' performance. A joint maximum likelihood (ML) estimator of the unknown labels and these parameters has been reported using the expectation-maximization (EM) algorithm [14]. As the EM algorithm does not guarantee convergence to the ML solution, recent works pursue alternative estimation methods. For binary classification, [15] assumes that annotators adhere to the "one-coin" model, meaning each annotator $m$ provides the correct (incorrect) label with probability $\delta_m$ $(1 - \delta_m)$; see also [16] when annotators do not label all the data, and [17] for an iterative method.

Recently, [18], [19] advocated a spectral decomposition technique of the second-order statistics of annotator responses for binary classification, that yields the reliability parameters of annotators, when class probabilities are unknown, while [20] introduced a minimax optimal algorithm that can infer annotator reliabilities. In the multiclass setting, [17] solves multiple binary classification problems. In addition, [21] and [22] utilize third-order moments and orthogonal tensor decomposition to estimate the unknown reliability parameters and then initialize the EM algorithm of [14].

This procedure however, can be numerically unstable, especially when the number of classes $K$ is large, and classes are unequally populated. Finally, all the methods mentioned in this section employ ML estimation, which implicitly assumes that the dataset is balanced, meaning classes are roughly equiprobable. Another interesting approach is presented in [23], where a joint moment matching and maximum likelihood optimization problem is solved.

The present work puts forth a novel scheme for *multiclass blind ensemble classification*, built upon simple concepts from probability and detection theory. It relies on a joint PARAFAC decomposition approach, which lends itself to a numerically stable algorithm. At the same time, our novel approach takes into account class prior probabilities to yield accurate estimates of class labels. Compared to our conference precursor in [24], here we do not require the prior probabilities to be known, and we present comprehensive numerical tests, along with a rigorous performance analysis.

### B. Canonical Polyadic Decomposition/PARAFAC

This subsection will outline tensor decompositions, which will be used in the following sections to derive the proposed scheme. Consider a 3-mode $I \times J \times L$ tensor $\underline{X}$, which can be described by a matrix in 3 different ways

$$\mathbf{X}^{(1)} := [\text{vec}(\underline{X}(1,:,:)), \ldots, \text{vec}(\underline{X}(I,:,:))] \quad (1a)$$

$$\mathbf{X}^{(2)} := [\text{vec}(\underline{X}(:,1,:)), \ldots, \text{vec}(\underline{X}(:,J,:))] \quad (1b)$$

$$\mathbf{X}^{(3)} := [\text{vec}(\underline{X}(:,:,1)), \ldots, \text{vec}(\underline{X}(:,:,L))] \quad (1c)$$

where $\mathbf{X}^{(1)}$ is of dimension $JL \times I$, $\mathbf{X}^{(2)}$ is $IL \times J$ and $\mathbf{X}^{(3)}$ is $IJ \times L$. Under the Canonical Polyadic Decomposition(CPD)/Parallel Factor Analysis (PARAFAC) model [25],

_$\underline{X}$ can be written as a sum of $R$ rank one tensors (a.k.a. *factors*)

$$\underline{X} = \sum_{r=1}^{R} \boldsymbol{a}_r \circ \boldsymbol{b}_r \circ \boldsymbol{c}_r \tag{2}$$

where $\boldsymbol{a}_r, \boldsymbol{b}_r, \boldsymbol{c}_r$ are $I \times 1, J \times 1$ and $L \times 1$ vectors, respectively. Letting $\mathbf{A} := [\boldsymbol{a}_1, \ldots, \boldsymbol{a}_R], \mathbf{B} := [\boldsymbol{b}_1, \ldots, \boldsymbol{b}_R]$, and $\mathbf{C} := [\boldsymbol{c}_1, \ldots, \boldsymbol{c}_R]$ be the so-called factor matrices of the CPD model, we write (2) compactly as

$$\underline{X} = [[\mathbf{A}, \mathbf{B}, \mathbf{C}]]_R \tag{3}$$

and (1) can be equivalently written as

$$\mathbf{X}^{(1)} = (\mathbf{C} \odot \mathbf{B}) \mathbf{A}^\top \tag{4a}$$

$$\mathbf{X}^{(2)} = (\mathbf{C} \odot \mathbf{A}) \mathbf{B}^\top \tag{4b}$$

$$\mathbf{X}^{(3)} = (\mathbf{B} \odot \mathbf{A}) \mathbf{C}^\top \tag{4c}$$

where we have used the fact that for matrices $\mathbf{A}, \mathbf{B}$ and a vector $\boldsymbol{c}$ of appropriate dimensions, it holds that $\text{vec}(\mathbf{A}\text{diag}(\boldsymbol{c})\mathbf{B}^\top) = (\mathbf{B} \odot \mathbf{A})\boldsymbol{c}$. By vectorizing $\mathbf{X}^{(3)}$, it is easy to show that the vectorization of the entire tensor will be of the form $\boldsymbol{x} := \text{vec}(\underline{X}) = \text{vec}(\mathbf{X}^{(3)}) = (\mathbf{C} \odot \mathbf{B} \odot \mathbf{A}) \mathbf{1}$.

Accordingly, vectorizing $\mathbf{X}^{(1)}$ or $\mathbf{X}^{(2)}$ produces different vectorizations of the entire tensor, where the order of factor matrices in the Khatri-Rao product is permuted. Recovery of the factor matrices $\mathbf{A}, \mathbf{B}$ and $\mathbf{C}$, can be done by solving the following non-convex optimization problem

$$[\hat{\mathbf{A}}, \hat{\mathbf{B}}, \hat{\mathbf{C}}] = \underset{\mathbf{A},\mathbf{B},\mathbf{C}}{\arg\min} \|\underline{X} - [[\mathbf{A}, \mathbf{B}, \mathbf{C}]]_R\|_F^2. \tag{5}$$

Similar to the matrix case, the Frobenius norm here can be defined as $\|\underline{X}\|_F := \sqrt{\sum_{i,j,l} \underline{X}(i,j,l)^2}$, and as (4) is just a rearrangement of the terms in $\underline{X}$, it holds that

$$\|\underline{X}\|_F = \|\mathbf{X}^{(1)}\|_F = \|\mathbf{X}^{(2)}\|_F = \|\mathbf{X}^{(3)}\|_F. \tag{6}$$

Typically, (5) is solved using alternating optimization (AO) or gradient descent [11]. Multiple off-the-shelf solvers are available for PARAFAC tensor decomposition; see e.g. [26], [27]. Furthermore, depending on extra properties of $\underline{X}$, constraints can be enforced on the factor matrices, such as nonnegativity and sparsity to name a few, which can be effectively handled by popular solvers such as AO-ADMM [28]. Under certain conditions, the factorization of $\underline{X}$ into $\mathbf{A}, \mathbf{B}$, and $\mathbf{C}$, is *essentially unique*, or *essentially identifiable*, that is $\hat{\mathbf{A}}, \hat{\mathbf{B}}$, and $\hat{\mathbf{C}}$ can be expressed as

$$\hat{\mathbf{A}} = \mathbf{A}\mathbf{P}\boldsymbol{\Lambda}_a, \quad \hat{\mathbf{B}} = \mathbf{B}\mathbf{P}\boldsymbol{\Lambda}_b, \quad \hat{\mathbf{C}} = \mathbf{C}\mathbf{P}\boldsymbol{\Lambda}_c \tag{7}$$

where $\mathbf{P}$ is a common permutation matrix, and $\boldsymbol{\Lambda}_a, \boldsymbol{\Lambda}_b, \boldsymbol{\Lambda}_c$ are diagonal scaling matrices such that $\boldsymbol{\Lambda}_a \boldsymbol{\Lambda}_b \boldsymbol{\Lambda}_c = \mathbf{I}$ [11]. For more details regarding the PARAFAC decomposition and tensors with more than 3 modes, interested readers are referred to the comprehensive tutorial in [11] and references therein.

## III. Unsupervised Ensemble Classification

Each annotator in our model has a fixed probability of deciding that a datum belongs to class $k'$, when presented with a datum of class $k$. Thus, each annotator $m$ can be characterized by a so called *confusion* matrix $\boldsymbol{\Gamma}_m$, whose $(k', k)$-th entry is

$$[\boldsymbol{\Gamma}_m]_{k'k} := \Gamma_m(k', k) = \Pr\left(f_m(X) = k'|Y = k\right). \tag{8}$$

The $K \times K$ matrix $\boldsymbol{\Gamma}_m$ has non-negative entries that obey the simplex constraint, since $\sum_{k'=1}^{K} \Pr\left(f_m(X) = k'|Y = k\right) = 1$, for $k = 1, \ldots, K$; hence, entries of each $\boldsymbol{\Gamma}_m$ column sum up to 1, that is, $\boldsymbol{\Gamma}_m^\top \mathbf{1} = \mathbf{1}$ and $\boldsymbol{\Gamma}_m \geq \mathbf{0}$. The confusion matrix showcases the statistical behavior of an annotator, as each column provides the annotator's probability of deciding the correct class, when presented with a datum from each class. Before proceeding, we adopt the following assumptions.

**As1.** Responses of different annotators per datum, are conditionally independent, given the ground-truth label $Y$ of the same datum $X$; that is,

$$\Pr\left(f_1(X) = k_1, \ldots, f_M(X) = k_M | Y = k\right)$$
$$= \prod_{m=1}^{M} \Pr\left(f_m(X) = k_m | Y = k\right)$$

**As2.** Most annotators are better than random; e.g., most have probability of correct detection exceeding 0.5 for $K = 2$.

Clearly, for annotators that are better than random, the largest elements of each column of their confusion matrix will be those on the diagonal of $\boldsymbol{\Gamma}_m$; that is

$$[\boldsymbol{\Gamma}_m]_{kk} \geq [\boldsymbol{\Gamma}_m]_{k'k}, \text{ for } k', k = 1, \ldots, K.$$

As1 suggests that annotators make decisions independently of each other, which is rather a standard assumption [14], [19], [22]. Likewise, As2 is another standard assumption, used to alleviate the inherent permutation ambiguity of the confusion matrix estimates provided by our algorithm. Note that As2 is slightly more relaxed than the corresponding assumption in [22], which splits annotators into 3 groups and requires most annotators in each group to be better than random.

### A. Maximum a posteriori label estimation

Given only annotator responses for all data, a straightforward approach to estimating their ground-truth labels is through a maximum a posteriori (MAP) classifier [29]. In particular, for datum $X$ the MAP classifier is

$$\hat{y}_{\text{MAP}}(X) = \arg\max_{k \in \{1,\ldots,K\}} L(X|k) \Pr(Y = k) \tag{9}$$

where $L(X|k) := \Pr\left(f_1(X) = k_1, \ldots, f_M(X) = k_M | Y = k\right)$ is the conditional likelihood of $X$. As annotators make independent decisions, it holds that $L(X|k) = \prod_{m=1}^{M} \Pr\left(f_m(X) = k_m | Y = k\right)$, and thus the MAP classifier can be rewritten as

$$\hat{y}_{\text{MAP}}(X) = \arg\max_{k \in \{1,\ldots,K\}} \log \pi_k + \sum_{m=1}^{M} \log(\Gamma_m(k_m, k)) \tag{10}$$

where $\pi_k := \Pr(Y = k)$. It is well known from detection theory [29] that the MAP classifier (10) minimizes the average probability of error $P_e$, given by

$$P_e = \sum_{k=1}^{K} \pi_k \Pr(\hat{y}_{\text{MAP}} = k' \neq k | Y = k). \quad (11)$$

If all classes are equiprobable, that is $\pi_k = 1/K$ for all $k = 1, \ldots, K$, then (10) reduces to the ML classifier. In order to obtain the MAP or ML classifier, $\{\mathbf{\Gamma}_m\}_{m=1}^{M}$ must be available, while in the MAP classifier case $\boldsymbol{\pi} := [\pi_1, \ldots, \pi_K]^\top$ is also required. Interestingly, the next section will illustrate that $\{\mathbf{\Gamma}_m\}_{m=1}^{M}$ and $\boldsymbol{\pi}$ show up in (and can thus be estimated from) the moments of annotator responses.

### B. Statistics of annotator responses

Consider each label represented by the annotators using the canonical $K \times 1$ vector $\boldsymbol{e}_k$, denoting the $k$-th column of the $K \times K$ identity matrix $\mathbf{I}$. Let $\mathbf{f}_m(X)$ denote the $m$-th annotator's response in vector format. Since $\mathbf{f}_m(X)$ is just a vector representation of $f_m(X)$, it holds that $\Pr(f_m(X) = k'|Y = k) \equiv \Pr(\mathbf{f}_m(X) = \boldsymbol{e}_{k'}|Y = k)$. With $\boldsymbol{\gamma}_{m,k}$ denoting the $k$-th column of $\mathbf{\Gamma}_m$, it thus holds that

$$\mathbb{E}[\mathbf{f}_m(X)|Y = k] = \sum_{k'=1}^{K} \boldsymbol{e}_{k'} \Pr(f_m(X) = k'|Y = k) \quad (12)$$
$$= \boldsymbol{\gamma}_{m,k}$$

where the first equality comes from the definition of conditional expectation, and the second one because $\boldsymbol{e}_k$'s are columns of $\mathbf{I}$. Using (12) and the law of total probability, the mean vector of responses from annotator $m$, is hence

$$\mathbb{E}[\mathbf{f}_m(X)] = \sum_{k=1}^{K} \mathbb{E}[\mathbf{f}_m(X)|Y = k] \Pr(Y = k) = \mathbf{\Gamma}_m \boldsymbol{\pi}. \quad (13)$$

Upon defining the diagonal matrix $\mathbf{\Pi} := \text{diag}(\boldsymbol{\pi})$, the $K \times K$ cross-correlation matrix between the responses of annotators $m$ and $m' \neq m$, can be expressed as

$$\mathbf{R}_{mm'} := \mathbb{E}[\mathbf{f}_m(X) \mathbf{f}_{m'}^\top(X)]$$
$$= \sum_{k=1}^{K} \mathbb{E}[\mathbf{f}_m(X)|Y = k] \mathbb{E}[\mathbf{f}_{m'}^\top(X)|Y = k] \Pr(Y = k)$$
$$= \mathbf{\Gamma}_m \text{diag}(\boldsymbol{\pi}) \mathbf{\Gamma}_{m'}^\top = \mathbf{\Gamma}_m \mathbf{\Pi} \mathbf{\Gamma}_{m'}^\top \quad (14)$$

where we successively relied on the law of total probability, As1, and (12). Consider now the $K \times K \times K$ cross-correlation tensor between the responses of annotators $m$, $m' \neq m$ and $m'' \neq m', m$, namely

$$\underline{\Psi}_{mm'm''} = \mathbb{E}[\mathbf{f}_m(X) \circ \mathbf{f}_{m'}(X) \circ \mathbf{f}_{m''}(X)]. \quad (15)$$

It can be shown that $\underline{\Psi}_{mm'm''}$ obeys a CPD/PARAFAC model [cf. Sec. II-B] with factor matrices $\mathbf{\Gamma}_m, \mathbf{\Gamma}_{m'}$ and $\mathbf{\Gamma}_{m''}$; that is,

$$\underline{\Psi}_{mm'm''} = \sum_{k=1}^{K} \pi_k \boldsymbol{\gamma}_{m,k} \circ \boldsymbol{\gamma}_{m',k} \circ \boldsymbol{\gamma}_{m'',k} \quad (16)$$
$$= [[\mathbf{\Gamma}_m \mathbf{\Pi}, \mathbf{\Gamma}_{m'}, \mathbf{\Gamma}_{m''}]]_K.$$

Note here that the diagonal matrix $\mathbf{\Pi}$ can multiply any of the factor matrices $\mathbf{\Gamma}_m, \mathbf{\Gamma}_{m'}$, or, $\mathbf{\Gamma}_{m''}$.

With $\mathbf{F}_m := [\mathbf{f}_m(x_1), \mathbf{f}_m(x_2), \ldots, \mathbf{f}_m(x_N)]$ the sample mean of the $m$-th annotator responses can be readily obtained as

$$\boldsymbol{\mu}_m = \frac{1}{N} \sum_{n=1}^{N} \mathbf{f}_m(x_n) = \frac{1}{N} \mathbf{F}_m \mathbf{1}. \quad (17)$$

Accordingly, the $K \times K$ sample cross-correlation $\mathbf{S}_{mm'}$ matrices between the responses of annotators $m$ and $m' \neq m$, are given by

$$\mathbf{S}_{mm'} = \frac{1}{N} \sum_{n=1}^{N} \mathbf{f}_m(\boldsymbol{x}_n) \mathbf{f}_{m'}^\top(\boldsymbol{x}_n) = \frac{1}{N} \mathbf{F}_m \mathbf{F}_{m'}^\top. \quad (18)$$

Lastly, the sample $K \times K \times K$ cross-correlation tensors $\underline{T}_{mm'm''}$ between the responses of annotators $m, m' \neq m$ and $m'' \neq m, m'$ are

$$\underline{T}_{mm'm''} = \frac{1}{N} \sum_{n=1}^{N} \mathbf{f}_m(\boldsymbol{x}_n) \circ \mathbf{f}_{m'}(\boldsymbol{x}_n) \circ \mathbf{f}_{m''}(\boldsymbol{x}_n) \quad (19)$$
$$= \frac{1}{N} \mathbf{F}_m \circ \mathbf{F}_{m'} \circ \mathbf{F}_{m''}.$$

Clearly, $\mathbf{S}_{mm'} = \mathbf{S}_{m'm}^\top$, $\mathbf{T}_{m'mm''}^{(2)} = \mathbf{T}_{m'm''m}^{(3)} = \mathbf{T}_{mm'm''}^{(1)}$. In addition, as $N$ increases, the law of large numbers (LLN) implies that, $\{\boldsymbol{\mu}_m\}, \{\mathbf{S}_{mm'}\}$, and $\{\underline{T}_{mm'm''}\}$ approach their ensemble counterparts in (13), (14), and (15).

Having available first-, second-, and third-order statistics of annotator responses, namely $\{\boldsymbol{\mu}_m\}_{m=1}^{M}, \{\mathbf{S}_{mm'}\}_{m,m'=1}^{M}$, and $\{\underline{T}_{mm'm''}\}_{m,m',m''=1}^{M}$, estimates of $\{\mathbf{\Gamma}_m\}_{m=1}^{M}$ and $\boldsymbol{\pi}$ can be readily extracted from them [cf. (13), (14), (15)]. This procedure corresponds to the method-of-moments estimation [30]. Upon obtaining $\{\hat{\mathbf{\Gamma}}_m\}_{m=1}^{M}$ and $\hat{\boldsymbol{\pi}}$, the MAP classifier of Sec. III-A can be subsequently employed to estimate the label for each datum. That is, for $n = 1, \ldots, N$,

$$\hat{y}_{\text{MAP}}(x_n) = \arg\max_{k \in \{1, \ldots, K\}} \log \hat{\pi}_k + \sum_{m=1}^{M} \log \hat{\Gamma}_m(f_m(x_n), k) \quad (20)$$

where $\hat{\Gamma}_m(k', k) = [\hat{\mathbf{\Gamma}}_m]_{k'k}$, and $\hat{\pi}_k = [\hat{\boldsymbol{\pi}}]_k$. The following section provides an algorithm to estimate these unknown quantities.

### C. Confusion matrix and prior probability estimation

To estimate the unknown confusion matrices and prior probabilities consider the following non-convex constrained optimization problem,

$$\min_{\substack{\boldsymbol{\pi} \\ \{\mathbf{\Gamma}_m\}_{m=1}^{M}}} h_N(\{\mathbf{\Gamma}_m\}_{m=1}^{M}, \boldsymbol{\pi}) \quad (21)$$

$$\text{s.to} \quad \mathbf{\Gamma}_m \geq \mathbf{0}, \quad \mathbf{\Gamma}_m^\top \mathbf{1} = \mathbf{1}, \quad m = 1, \ldots, M$$
$$\boldsymbol{\pi} \geq \mathbf{0}, \quad \boldsymbol{\pi}^\top \mathbf{1} = 1$$

where

**Algorithm 1** Confusion matrix and prior probability estimation algorithm

**Input:** Annotator responses $\{\mathbf{F}_m\}_{m=1}^M$, $\lambda > 0$, $\nu > 0$; maximum number of iterations $I \in \mathbb{Z}_+$
**Output:** Estimates of $\{\hat{\mathbf{\Gamma}}_m\}_{m=1}^M$ and $\hat{\boldsymbol{\pi}}$
1: Compute $\{\boldsymbol{\mu}_m\}, \{\mathbf{S}_{mm'}\}, \{\underline{T}_{mm'm''}\}$ using (17), (18), and (19).
2: Initialize $\{\mathbf{\Gamma}_m\}$ and $\boldsymbol{\pi}$ randomly.
3: **do**
4:   **for** $m = 1, \ldots, M$ **do**
5:     Update $\mathbf{\Gamma}_m$ using (23)
6:     $\mathbf{\Gamma}_m^{(\text{prev})} \leftarrow \mathbf{\Gamma}_m$
7:   **end for**
8:   Update $\boldsymbol{\pi}$ using (22)
9:   $\boldsymbol{\pi}^{(\text{prev})} \leftarrow \boldsymbol{\pi}$
10:   $i \leftarrow i + 1$
11: **while** not converged and $i < I_T$
12: Find permutation matrix $\hat{\mathbf{P}}$, such that the majority of $\{\hat{\mathbf{\Gamma}}_m\hat{\mathbf{P}}\}_{m=1}^M$ satisfy As2.

**Algorithm 2** Unsupervised multiclass ensemble classification

**Input:** Annotator responses $\{\mathbf{F}_m\}_{m=1}^M$
**Output:** Estimates of data labels $\{\hat{y}_n\}_{n=1}^N$
1: Find estimates $\{\hat{\mathbf{\Gamma}}_m\}_{m=1}^M$ and $\hat{\boldsymbol{\pi}}$ using Alg. 1
2: **for** $n = 1, \ldots, N$ **do**
3:   Estimate label $y_n$ using (20).
4: **end for**

$$h_N(\{\mathbf{\Gamma}_m\}, \boldsymbol{\pi}) := \frac{1}{2} \sum_{m=1}^M \|\boldsymbol{\mu}_m - \mathbf{\Gamma}_m\boldsymbol{\pi}\|_2^2$$
$$+ \frac{1}{2} \sum_{\substack{m=1 \\ m'>m}}^M \|\mathbf{S}_{mm'} - \mathbf{\Gamma}_m\mathbf{\Pi}\mathbf{\Gamma}_{m'}^\top\|_F^2$$
$$+ \frac{1}{2} \sum_{\substack{m=1 \\ m'>m \\ m''>m'}}^M \|\underline{T}_{mm'm''} - [[\mathbf{\Gamma}_m\mathbf{\Pi}, \mathbf{\Gamma}_{m'}, \mathbf{\Gamma}_{m''}]]_K\|_F^2$$

and the subscript $N$ in $h_N$ denotes the number of data used to obtain annotator statistics. Collect the set of constraints per matrix to the convex set $\mathcal{C} := \{\mathbf{\Gamma} \in \mathbb{R}^{K \times K} : \mathbf{\Gamma} \geq \mathbf{0}, \mathbf{\Gamma}^\top\mathbf{1} = \mathbf{1}\}$, where essentially each column lies on a probability simplex, and let $\mathcal{C}_p := \{\boldsymbol{u} \in \mathbb{R}^K : \boldsymbol{u} \geq \mathbf{0}, \boldsymbol{u}^\top\mathbf{1} = 1\}$ denote the constraint set for $\boldsymbol{\pi}$.

As (21) is a non-convex problem, alternating optimization will be employed to solve it. Specifically the alternating optimization-alternating direction method of multipliers (AO-ADMM) will be employed; see [28], and also [12] where a similar formulation appears. Under the AO-ADMM paradigm, $h_N$ is minimized per block of unknown variables $\{\mathbf{\Gamma}_m\}$ or $\boldsymbol{\pi}$ while the other blocks remain fixed, as in block coordinate descent schemes. Solving for one block of variables with the remaining fixed is a convex constrained optimization problem under convex $\mathcal{C}$ and $\mathcal{C}_p$ constraint sets. These optimization problems are pretty standard and several solvers are available, including proximal splitting methods, projected gradient descent or ADMM [31]–[34]. Here, the solver of choice for each block of variables will be ADMM.

The update for $\boldsymbol{\pi}$ involves minimizing $h_N$ with $\{\mathbf{\Gamma}_m\}_{m=1}^M$ fixed. Specifically, the following problem is solved

$$\min_{\boldsymbol{\pi} \in \mathcal{C}_p} \quad g_{N,\boldsymbol{\pi}}(\boldsymbol{\pi}) \tag{22}$$

where

$$g_{N,\boldsymbol{\pi}}(\boldsymbol{\pi}) := \frac{1}{2} \sum_{m=1}^M \|\boldsymbol{\mu}_m - \mathbf{\Gamma}_m\boldsymbol{\pi}\|_2^2 + \frac{\nu}{2}\|\boldsymbol{\pi} - \boldsymbol{\pi}^{(\text{prev})}\|_2^2$$
$$+ \frac{1}{2} \sum_{\substack{m=1 \\ m'>m}}^M \|\boldsymbol{s}_{mm'} - (\mathbf{\Gamma}_{m'} \odot \mathbf{\Gamma}_m)\boldsymbol{\pi}\|_2^2$$
$$+ \frac{1}{2} \sum_{\substack{m=1 \\ m'>m \\ m''>m'}}^M \|\boldsymbol{t}_{mm'm''} - (\mathbf{\Gamma}_{m''} \odot \mathbf{\Gamma}_{m'} \odot \mathbf{\Gamma}_m)\boldsymbol{\pi}\|_2^2$$

$\boldsymbol{s}_{mm'} = \text{vec}(\mathbf{S}_{mm'})$, $\boldsymbol{t}_{mm'm''} = \text{vec}(\mathbf{T}^{(3)}_{mm'm''})$ [cf. (4)], $\nu$ is a positive scalar, and we have used $\text{vec}(\mathbf{\Gamma}_m\text{diag}(\boldsymbol{\pi})\mathbf{\Gamma}_{m'}^\top) = (\mathbf{\Gamma}_{m'} \odot \mathbf{\Gamma}_m)\boldsymbol{\pi}$ and $\text{vec}([[\mathbf{\Gamma}_m\text{diag}(\boldsymbol{\pi}), \mathbf{\Gamma}_{m'}, \mathbf{\Gamma}_{m''}]]_K) = (\mathbf{\Gamma}_{m''} \odot \mathbf{\Gamma}_{m'} \odot \mathbf{\Gamma}_m)\boldsymbol{\pi}$. Note that $g_{N,\boldsymbol{\pi}}$ contains all of the terms in $h_N$ along with $(\nu/2)\|\boldsymbol{\pi} - \boldsymbol{\pi}^{(\text{prev})}\|_2^2$, which is included to ensure convergence of the AO-ADMM iterations to a stationary point of (21) [28], [35]. Here, $\boldsymbol{\pi}^{(\text{prev})}$ denotes the estimate of $\boldsymbol{\pi}$ obtained by the previous solutions of (22).

Accordingly per $\mathbf{\Gamma}_m$, the following subproblem is solved with $\{\mathbf{\Gamma}_{m'}\}_{m' \neq m}^M$ and $\boldsymbol{\pi}$ fixed

$$\min_{\mathbf{\Gamma}_m \in \mathcal{C}} \quad g_{N,m}(\mathbf{\Gamma}_m) \tag{23}$$

where

$$g_{N,m}(\mathbf{\Gamma}_m) := \frac{1}{2}\|\boldsymbol{\mu}_m - \mathbf{\Gamma}_m\boldsymbol{\pi}\|_2^2 + \frac{\nu}{2}\|\mathbf{\Gamma}_m - \mathbf{\Gamma}_m^{(\text{prev})}\|_F^2$$
$$+ \frac{1}{2} \sum_{m' \neq m}^M \|\mathbf{S}_{m'm} - \mathbf{\Gamma}_{m'}\mathbf{\Pi}\mathbf{\Gamma}_m^\top\|_F^2$$
$$+ \frac{1}{2} \sum_{\substack{m''>m \\ m''>m'}}^M \|\mathbf{T}^{(1)}_{mm'm''} - (\mathbf{\Gamma}_{m''} \odot \mathbf{\Gamma}_{m'})\mathbf{\Pi}\mathbf{\Gamma}_m^\top\|_F^2$$

$\mathbf{T}^{(1)}_{mm'm''} = [\text{vec}(\underline{T}(1,:,:)), \ldots, \text{vec}(\underline{T}(K,:,:))]$, $\mathbf{\Gamma}_m^{(\text{prev})}$ denotes the estimate of $\mathbf{\Gamma}_m$ obtained by the previous solution of (23), $\nu$ is a positive scalar, and we have used (6). Here, $g_{N,m}$ contains all the terms of $h_N$ that involve $\mathbf{\Gamma}_m$ with the additional term $(\nu/2)\|\mathbf{\Gamma}_m - \mathbf{\Gamma}_m^{(\text{prev})}\|_F^2$, which ensures convergence of the AO-ADMM iterations.

Detailed derivations of the ADMM iterations for solving (23) and (22) are provided in Appendix A, while the AO-ADMM is summarized in Alg. 1. The computational complexity of the entire AO-ADMM scheme is approximately $\mathcal{O}(I_T M^3 K^4)$, where $I_T$ is the number of required iterations until convergence (see Appendix A-C). The entire unsupervised ensemble classification procedure is listed in Alg. 2.

*D. Convergence and identifiability*

Convergence of the entire AO-ADMM scheme for (21), follows readily from results in [28, Prop. 1], stated next for our setup.

**Proposition 1.** *[28, Prop. 1] Alg. 1 for $M \geq 3$, and $\nu > 0$ converges to a stationary point of (21).*

Having established the convergence of Alg. 1 to a stationary point of (21) using Prop. 1, the suitability of the estimates provided by Alg. 1 for the ensemble classification task needs to be assessed. As (21) involves joint tensor decompositions, under certain conditions the solutions $\{\hat{\boldsymbol{\Gamma}}_m\}, \hat{\boldsymbol{\pi}}$ of (21) will be, similar to the PARAFAC decomposition of Sec. II-B, *essentially unique*.

Thus, in order to assess the suitability of the estimates provided by Alg. 1 the conditions under which the model employed in (21) is identifiable have to be established. Luckily, identifiability claims for the present problem can be easily derived from recent results in joint PARAFAC factorization [12], [36].

**Lemma 1.** *Let $\{\boldsymbol{\Gamma}_m^*\}, \boldsymbol{\pi}^*$ be the optimal solutions of (21), and $\{\hat{\boldsymbol{\Gamma}}_m\}, \hat{\boldsymbol{\pi}}$ the estimates provided by Alg. 1. If at least three $\{\boldsymbol{\Gamma}_m\}_{m=1}^M$ have full column rank, there exists a permutation matrix $\hat{\mathbf{P}}$ such that*

$$\hat{\boldsymbol{\Gamma}}_m \hat{\mathbf{P}} = \boldsymbol{\Gamma}_m^*, \quad m = 1, \ldots, M, \quad \hat{\mathbf{P}}^\top \hat{\boldsymbol{\pi}} = \boldsymbol{\pi}^*.$$

Lemma 1 essentially requires that at least three annotators respond differently to different classes, that is no two columns of at least three confusion matrices are colinear. Possibly more relaxed identifiability conditions could be derived using techniques mentioned in [36].

Unlike the tensor decomposition mentioned in Sec. II-B, here we have no scaling ambiguity on the confusion matrices or prior probabilities. This is important because there are infinite scalings, but finite permutation matrices since $K$ is finite. Under As2, $\hat{\mathbf{P}}$ can be easily obtained since the largest elements of each column of a confusion matrix must lie on the diagonal for the majority of annotators. Each $\hat{\boldsymbol{\Gamma}}_m$ can be multiplied by a permutation matrix $\hat{\mathbf{P}}_m$, such that the largest elements are located on the diagonal. The final $\hat{\mathbf{P}}$ can be derived as the most commonly occurring permutation matrix out of $\{\hat{\mathbf{P}}_m\}_{m=1}^M$.

**Remark 1.** While we relied on statistics of annotator responses up to order three, higher-order statistics can also be employed. Higher-order moments however, will increase the complexity of the algorithm, as well as the number of data required to obtain reliable (low-variance) estimates.

**Remark 2.** Estimates of annotator confusion matrices $\{\hat{\boldsymbol{\Gamma}}_m\}$ and data labels $\{\hat{y}_n\}$, provided by Alg. 2, can be used to initialize the EM algorithm of [14].

**Remark 3.** The orthogonal tensor decomposition used by [21], [22] is a special case of the PARAFAC decomposition employed in this work.

**Remark 4.** When $\boldsymbol{\pi}$ is known, (22) can be skipped, and correspondingly steps 8 and 9 of Alg. 1.

*E. Reducing complexity*

When $K$ and $M$ are large Alg. 1 may require long computational time to converge. Our idea in this case is to split the annotators into $L$ groups, and solve (21) for each group. For simplicity of exposition, consider non-overlapping groups, each with $M_\ell \geq 3$ annotators ($\sum_{\ell=1}^L M_\ell = M$). Let $\boldsymbol{\mu}_m^{(\ell)}, \mathbf{S}_{mm'}^{(\ell)}$ and $\underline{T}_{mm'm''}^{(\ell)}$ denote the sample statistics for annotators in group $\ell$, and $\{\boldsymbol{\Gamma}_m^{(\ell)}\}_{m=1}^{M_\ell}$ the confusion matrices in group $\ell$.

For each group $\ell \in \{1, \ldots, L\}$ confusion matrices $\{\hat{\boldsymbol{\Gamma}}_m^{(\ell)}\}_{m=1}^{M_\ell}$ and prior probabilities $\boldsymbol{\pi}^{(\ell)}$ are estimated by solving a smaller version of (21), namely

$$\min_{\substack{\boldsymbol{\pi}^{(\ell)} \\ \{\boldsymbol{\Gamma}_m^{(\ell)}\}_{m=1}^{M_\ell}}} h_N^{(\ell)}(\{\boldsymbol{\Gamma}_m^{(\ell)}\}_{m=1}^M, \boldsymbol{\pi}^{(\ell)}) \quad (24)$$

$$\text{s.to} \quad \boldsymbol{\Gamma}_m^{(\ell)} \geq \mathbf{0}, \quad \mathbf{1}^\top \boldsymbol{\Gamma}_m^{(\ell)} = \mathbf{1}^\top, \quad m = 1, \ldots, M_\ell$$
$$\boldsymbol{\pi}^{(\ell)} \geq \mathbf{0}, \quad \mathbf{1}^\top \boldsymbol{\pi}^{(\ell)} = 1$$

where

$$h_N^{(\ell)}(\{\boldsymbol{\Gamma}_m\}, \boldsymbol{\pi}) := \frac{1}{2} \sum_{m=1}^{M_\ell} \|\boldsymbol{\mu}_m^{(\ell)} - \boldsymbol{\Gamma}_m \boldsymbol{\pi}\|_2^2$$
$$+ \frac{1}{2} \sum_{\substack{m=1 \\ m'>m}}^{M_\ell} \|\mathbf{S}_{mm'}^{(\ell)} - \boldsymbol{\Gamma}_m \boldsymbol{\Pi} \boldsymbol{\Gamma}_{m'}^\top\|_F^2$$
$$+ \frac{1}{2} \sum_{\substack{m=1 \\ m'>m \\ m''>m'}}^{M} \|\underline{T}_{mm'm''}^{(\ell)} - [[\boldsymbol{\Gamma}_m \boldsymbol{\Pi}, \boldsymbol{\Gamma}_{m'}, \boldsymbol{\Gamma}_{m''}]]_K\|_F^2.$$

Upon solving (24) for all $L$ groups, estimates of $\{\boldsymbol{\Gamma}_m\}_{m=1}^M$ are readily obtained, since we have assumed non-overlapping groups. A final estimate of the prior probabilities $\boldsymbol{\pi}$ can be obtained by averaging the $L$ estimates $\{\boldsymbol{\pi}^\ell\}_{\ell=1}^L$.

As (24) incurs a complexity of $\mathcal{O}(IM_\ell^3 K^3)$, the worst-case complexity of this approach is $\mathcal{O}(I_M K^3 \sum_{\ell=1}^L M_\ell^3)$, where $I_M$ is the largest number of iterations required to converge among all $L$ groups. Since $M^3 = (\sum_{\ell=1}^L M_\ell)^3 > \sum_{\ell=1}^L M_\ell^3$ this approach reduces the computational and memory overhead significantly compared to Alg. 1. Note however, that this method is expected to perform well when As1 and As2, as well as the conditions outlined in Lemma 1 are satisfied for all $L$ groups of annotators, and $N$ is sufficiently large. The effectiveness of this complexity reduction scheme is tested in Sec. V.

*F. Application to crowdsourcing*

While crowdsourced classification is a task related to ensemble classification, it presents additional challenges. So far it has been implicitly assumed that all annotators provide labels for all $\{x_n\}_{n=1}^N$. In the crowdsourcing setup however, an annotator $m$ could provide labels just for a subset of $N_m < N$ data.

Next, we outline a computationally attractive approach, that takes into account only the available annotator responses. If an annotator $m$ does not provide a label for a datum, his/her response is $f_m(x) = 0$ or $\mathbf{f}_m(x) = \mathbf{0}$ in vector format. Let $J_m(x_n)$ be an indicator function that takes the value 1 when annotator $m$ provides a label for $x_n$, and 0 when $f_m(x_n) = 0$. To account for such cases, the annotator sample statistics become

$$\boldsymbol{\mu}_m = \frac{1}{\sum_{n=1}^N J_m(x_n)} \sum_{n=1}^N J_m(x_n)\mathbf{f}_m(x_n) \quad (25a)$$

$$\mathbf{S}_{mm'} = \frac{\sum_{n=1}^N J_m(x_n)J_{m'}(x_n)\mathbf{f}_m(\boldsymbol{x}_n)\mathbf{f}_{m'}^\top(\boldsymbol{x}_n)}{\sum_{n=1}^N J_m(x_n)J_{m'}(x_n)} \quad (25b)$$

$$\underline{T}_{mm'm''} \quad (25c)$$
$$= \frac{\sum_n J_m(x_n)J_{m'}(x_n)J_{m''}(x_n)\mathbf{f}_m(\boldsymbol{x}_n) \circ \mathbf{f}_{m'}(\boldsymbol{x}_n) \circ \mathbf{f}_{m''}(\boldsymbol{x}_n)}{\sum_{n=1}^N J_m(x_n)J_{m'}(x_n)J_{m''}(x_n)}.$$

Upon computing the modified sample statistics of (25), we can obtain estimates of the confusion matrices and prior probabilities in the crowdsourcing setup, via Alg. 1. Finally, the MAP classifier in (20) has to be modified as follows

$$\hat{y}_{\text{MAP}}(x) = \underset{k \in \{1,\ldots,K\}}{\arg\max} \log \hat{\pi}_k + \sum_{m=1}^M J_m(x) \log \hat{\Gamma}_m(f_m(x), k) \quad (26)$$

to take into account only the available annotator responses for each $x$.

Having completed the algorithmic aspects of our approach, we proceed with performance analysis.

## IV. PERFORMANCE ANALYSIS

In this section, performance of the proposed method will be quantified analytically. First, the consistency of the estimates provided by Alg. 1 as $N \to \infty$ will be established, followed by a performance analysis for the MAP classifier of Sec. III-A.

### A. Consistency of Alg. 1 estimates

As $N \to \infty$, the sample statistics in (17), (18), and (19) approach their ensemble counterparts, and we end up with the following optimization problem for extracting annotator confusion matrices and prior probabilities

$$\min_{\substack{\boldsymbol{\pi} \\ \{\boldsymbol{\Gamma}_m\}_{m=1}^M}} h_\infty(\{\boldsymbol{\Gamma}_m\}_{m=1}^M, \boldsymbol{\pi}) \quad (27)$$
$$\text{s.to} \quad \boldsymbol{\Gamma}_m \in \mathcal{C}, \quad m = 1,\ldots,M, \quad \boldsymbol{\pi} \in \mathcal{C}_p.$$

Clearly, the optimal solutions to (27) are the true confusion matrices and prior probabilities. As $N$ increases, it is desirable to show that the solutions obtained from Alg. 1 converge to the true confusion matrices and prior probabilities. To this end, techniques from statistical learning theory and stochastic optimization will be employed [37], [38]. Specifically, we will establish the uniform convergence of $h_N$ to $h_\infty$, which implies the consistency of the solutions. Define the distance between two sets $\mathcal{A}, \mathcal{B} \subseteq \mathbb{R}^q$, for some $q > 0$, as $D(\mathcal{A}, \mathcal{B}) = \sup_{x \in \mathcal{A}}\{\inf_{y \in \mathcal{B}} \|x - y\|_2\}$.

The following theorem shows that as $N$ increases, the solutions of (21) approach those of (27).

**Theorem 1.** *If $\mathcal{S}_*$ and $\mathcal{S}_N$ denote the sets of solutions of problems (27) and (21), respectively, then $D(\mathcal{S}_N, \mathcal{S}_*) \to 0$, as $N \to \infty$ almost surely.*

Under As2 and the conditions outlined in Lemma 1, Alg. 1 can recover the true solutions of (21) or (27). Then, by Thm. 1 we know that as $N \to \infty$ the solutions of (21) converge to the solutions of (27), which together with the result of Lemma 1 implies the statistical consistency of the solutions of Alg. 1. As a result, the estimates $\{\hat{\boldsymbol{\Gamma}}_m\}_{m=1}^M$, and $\hat{\boldsymbol{\pi}}$ from Alg. 1 will converge to their true values w.p. 1 as $N \to \infty$.

### B. MAP classifier performance

With consistency of the confusion matrix and prior probability estimates established, the performance of the final component of the proposed algorithm has to be studied. The behavior of the MAP classifier of Sec. III-A can be quantified in terms of its average probability of error

$$\text{P}_\text{e} = \sum_{k=1}^K \Pr(\hat{y}_\text{MAP} = k' \neq k | Y = k) \Pr(Y = k)$$

Here, a well-known asymptotic result for distributed binary detection under the MAP detector [6] is extended to the multiclass case.

**Theorem 2.** *Under As1, and given $\{\boldsymbol{\Gamma}_m\}_{m=1}^M$ and $\boldsymbol{\pi}$, there exist constants $\alpha > 0, \beta > 0$ such that the MAP classifier of Sec. III-A satisfies*

$$\text{P}_\text{e} \leq \alpha e^{-\beta M}.$$

In words, Theorem 2 suggests that when accurate estimates of $\{\boldsymbol{\Gamma}_m\}_{m=1}^M$ and $\boldsymbol{\pi}$ are available, the error rate decreases at an exponential rate with the number of annotators $M$.

In order to validate our theoretical results and evaluate the performance of the proposed scheme, the following section presents numerical tests with synthetic and real data.

## V. NUMERICAL TESTS

For $K \geq 2$, Alg. 2, using both MAP and ML criteria in step 3, (denoted as *Alg. 2 MAP* and *Alg. 2 ML* respectively) is compared to majority voting, the algorithm of [17] (denoted as *KOS*), and the EM algorithm initialized both with majority voting and with the spectral method of [22] (denoted as *EM + MV* and *EM + Spectral*, respectively). For $K = 2$, Alg. 2 is also compared to the binary ensemble learning methods of [19], [20] and [16], denoted as *SML*, *TE* and *EigenRatio*, respectively. For synthetic data, the performance of "oracle" estimators, that is MAP/ML classifiers with true confusion matrices of the annotators, and the true class priors, is also evaluated for benchmarking purposes. The metric utilized in all experiments is the classification error rate (ER), defined as the percentage of misclassified data,

$$ER = \frac{\# \text{ of misclassified data}}{N} \times 100\%,$$

where $ER = 100\%$ indicates that all $N$ data have been misclassified, and $ER = 0\%$ indicates perfect classification accuracy. For synthetic data, the average confusion matrix and prior probability estimation error is also evaluated

$$\bar{\varepsilon}_{CM} := \frac{1}{M} \sum_{m=1}^{M} \frac{\|\mathbf{\Gamma}_m - \hat{\mathbf{\Gamma}}_m\|_1}{\|\mathbf{\Gamma}_m\|_1} = \frac{1}{M} \sum_{m=1}^{M} \|\mathbf{\Gamma}_m - \hat{\mathbf{\Gamma}}_m\|_1$$

$$\bar{\varepsilon}_{\boldsymbol{\pi}} := \|\boldsymbol{\pi} - \hat{\boldsymbol{\pi}}\|_1.$$

All results represent averages over 10 independent Monte Carlo runs, using MATLAB [39]. In all experiments, the parameters $\lambda$ and $\nu$ of Alg. 1 are set as suggested in [28], [35]. Vertical lines in some figures indicate standard deviation. For some experiments, classification times (in seconds) required by the ensemble algorithms are also reported. Note that classification times for majority voting and oracle estimators are not reported as the time required by these methods is negligible compared to the rest of the algorithms.

### A. Synthetic data

For the synthetic data tests, $N$ ground-truth labels $\{y_n\}_{n=1}^{N}$, each corresponding to one out of $K$ possible classes, were generated i.i.d. at random according to $\boldsymbol{\pi}$, that is $y_n \sim \boldsymbol{\pi}$, for $n = 1, \ldots, N$. Afterwards, $\{\mathbf{\Gamma}_m\}_{m=1}^{M}$ were generated at random, such that $\mathbf{\Gamma}_m \in \mathcal{C}$, for all $m = 1, \ldots, M$, and annotators are better than random, as per As2. Then annotators' responses were generated as follows: if $y_n = k$, then the response of annotator $m$ will be generated randomly according to the $k$-th column of its confusion matrix, $\boldsymbol{\gamma}_{m,k}$ [cf. Sec. II], that is $f_m(x_n) \sim \boldsymbol{\gamma}_{m,k}$.

Tab. I lists the classification ER of different algorithms, for a synthetic dataset with $K = 2$ classes with prior probabilities $\boldsymbol{\pi} = [0.9003, 0.0997]^\top$, and $M = 10$ annotators. Tab. II lists the results for a similar experiment, with $K = 2$ classes, priors $\boldsymbol{\pi} = [0.5856, 0.4144]^\top$, and $M = 10$ annotators, while Tab. III shows the clustering time required by all algorithms. Note that when the class probabilities are similar, the ML and MAP classifiers perform comparably as expected. Furthermore, majority voting gives good results for a reduced number of instances $N$. Fig. 2 depicts the average estimation errors for the confusion matrices and prior probabilities in the two aforementioned experiments. Clearly, as $N$ increases, the proposed classifiers approach the performance of the oracle ones, and as suggested by Thm. 1, the estimation error for the confusion matrices and prior probabilities approaches 0.

The next synthetic data experiment investigates how the proposed method performs when presented with multiclass data. Furthermore, to showcase that accurate estimation of $\boldsymbol{\pi}$ is beneficial, we also compare against Alg. 2 with $\boldsymbol{\pi}$ fixed to the uniform distribution, i.e. $\boldsymbol{\pi} = \mathbf{1}/K$ (denoted as *Alg. 2 - fixed $\boldsymbol{\pi}$.*) Fig. 3 shows the simulation results for a synthetic dataset with $K = 5$ classes, prior probabilities $\boldsymbol{\pi} = [0.2404, 0.2679, 0.0731, 0.1950, 0.2236]^\top$, and $M = 10$ annotators, while Fig. 4 shows the simulation results for a synthetic dataset with $K = 7$ classes, priors $\boldsymbol{\pi} = [0.2347, 0.0230, 0.0705, 0.1477, 0.2659, 0.0043, 0.2539]^\top$ and $M = 10$ annotators. Tabs. IV and V show classification times for the $K = 5$ and $K = 7$ experiments, respectively. Fig. 5 shows the average estimation errors for the confusion matrices and prior probabilities in the two aforementioned multiclass experiments. Note that for $K = 5$ for small values of $N$ and $K = 7$ the EM+Spectral approach of [22] suffers from numerical issues during the tensor whitening procedure, which explains its worst classification ER and slow runtimes. Here, the proposed approaches exhibit similar behavior to the binary case, as expected from Thm. 1; *as the number of data increases, their performance approaches the clairvoyant "oracle" estimators, and the estimation accuracy of the confusion matrices and prior probabilities increases*. In addition, our methods outperform the competing alternatives for almost all values of $N$. Here we also see that running Alg. 2 with fixed $\boldsymbol{\pi} = 1/K$ produces lower quality estimates than Alg. 2 that solves for $\boldsymbol{\pi}$. Specifically, Alg. 2 with fixed $\boldsymbol{\pi}$ performs similarly to the EM algorithm when initialized with majority voting.

Next, we evaluate how the number of annotators $M$ affects the classification ER, for fixed $N = 10^6$. Fig. 6 depicts an experiment for $K = 3$ classes with priors $\boldsymbol{\pi} = [0.2318, 0.4713, 0.2969]^\top$, while Fig. 7 shows an experiment for $K = 5$ classes with priors $\boldsymbol{\pi} = [0.3596, 0.1553, 0.1229, 0.3258, 0.0364]^\top$. Tabs. VI and VII list classification times for the $K = 3$ and $K = 5$ experiments, respectively. Fig. 8 plots the results of an experiment with $K = 5$ classes with the same priors as those in Fig. 7 and $N = 5,000$ data, for varying number of annotators. The average estimation error for the confusion matrices and prior probabilities, for the aforementioned tests, is shown in Fig. 9. As expected from Thm. 2, *the classification ER decreases as the number of annotators increase*, for all methods considered. In addition, our proposed algorithm outperforms the competing alternatives for all values of $M$. Furthermore, the results of Fig. 8 indicate that *when the number of data is small, increasing the number of annotators provides a boost to the classification performance*. Fig. 9 shows another interesting feature: *as the number of annotators increases the estimation accuracy of $\{\mathbf{\Gamma}_m\}$ and $\boldsymbol{\pi}$ also increases*.

The following experiment evaluates the effectiveness of the complexity reduction scheme of Sec. III-E, for a dataset with $M = 30$ annotators with $K = 3$ classes with priors $\boldsymbol{\pi} = [0.3096, 0.3416, 0.3488]^\top$, and a varying number of data $N$. Annotators are split into $L = \{1, 2, 4, 5\}$ non-overlapping groups. Fig. 10 shows the classifcation ER and time (in seconds) required for the ensemble classification task, for different group sizes. When $N$ is large we observe similar ER for all $L$, however larger number of groups require significantly less time than $L = 1$.

In all aforementioned experiments, all annotators were generated to be better than random. The next experiment, investigates the effect of *adversarial annotators*, that is annotators for who the largest values of the confusion matrix are not located on its diagonal. Let $\alpha$ denote the percentage of adversarial annotators. Fig. 11 shows the classification ER on a synthetic dataset with $K = 3$, $N = 10^6$, $\boldsymbol{\pi} = [0.31, 0.34, 0.35]^\top$ and $M = 10$ annotators, for varying $\alpha$. While all approaches, with the exception of majority voting, seem to be robust to a small number of adversarial annotators, Alg. 2 can handle values of $\alpha$ of up to $50\%$, which speaks for the potential of the novel

| Algorithm | $N=100$ | $N=1000$ | $N=10^4$ | $N=10^5$ |
|---|---|---|---|---|
| Majority Voting | 6.3 | 7.08 | 7.04 | 7.13 |
| KOS | 27.70 | 33.33 | 32.21 | 32.53 |
| EigenRatio | 6.30 | 5.75 | 5.69 | 5.64 |
| TE | 4.20 | 4.91 | 4.61 | 4.67 |
| SML | 15.80 | 11.38 | 11.82 | 12.26 |
| EM + MV | 21.2 | 27.67 | 26.50 | 27.01 |
| EM + Spectral | 17.7 | 27.72 | 26.50 | 27.01 |
| Alg. 2 ML | 6.30 | **2.70** | **1.97** | **1.87** |
| Alg. 2 MAP | **2.40** | **1.40** | **1.13** | **1.11** |
| Oracle ML | 1.6 | 2.05 | 1.81 | 1.86 |
| Oracle MAP | 1.1 | 1.31 | 1.11 | 1.11 |

TABLE I
CLASSIFICATION ER FOR A SYNTHETIC DATASET WITH $K=2$, PRIOR PROBABILITIES $\boldsymbol{\pi} = [0.9003, 0.0997]^\top$ AND $M=10$ ANNOTATORS.

| Algorithm | $N=100$ | $N=1000$ | $N=10^4$ | $N=10^5$ |
|---|---|---|---|---|
| Majority Voting | 8.10 | 8.27 | 8.27 | 8.19 |
| KOS | 8.30 | 6.46 | 6.65 | 6.58 |
| EigenRatio | 7.40 | 6.35 | 6.39 | 6.21 |
| TE | 10.20 | 6.04 | 6.35 | 6.20 |
| SML | 13.10 | 8.47 | **4.66** | **4.61** |
| EM + MV | 6.60 | 5.15 | 4.93 | 4.87 |
| EM + Spectral | 6.60 | 5.15 | 4.93 | 4.87 |
| Alg. 2 ML | **6.50** | **4.86** | **4.66** | **4.61** |
| Alg. 2 MAP | **6.20** | **4.85** | **4.59** | **4.51** |
| Oracle ML | 4.10 | 4.86 | 4.66 | 4.61 |
| Oracle MAP | 3.90 | 4.81 | 4.58 | 4.50 |

TABLE II
CLASSIFICATION ER FOR A SYNTHETIC DATASET WITH $K=2$, PRIOR PROBABILITIES $\boldsymbol{\pi} = [0.5856, 0.4144]^\top$ AND $M=10$ ANNOTATORS.

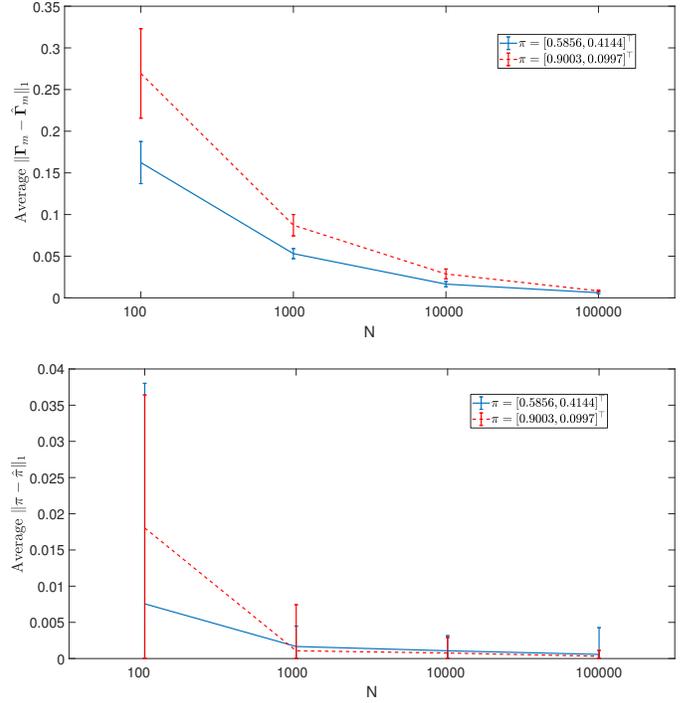

Fig. 2. Average estimation errors of confusion matrices (top); and prior probabilities (bottom), for two synthetic datasets with $K=2$ and $M=10$ annotators

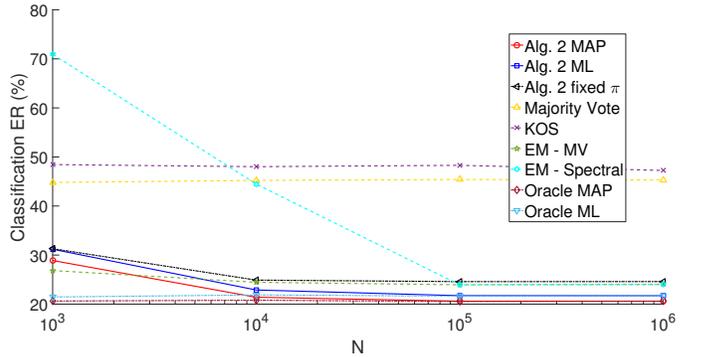

Fig. 3. Classification ER for a synthetic dataset with $K=5$ classes, priors $\boldsymbol{\pi} = [0.2404, 0.2679, 0.0731, 0.1950, 0.2236]^\top$ and $M=10$ annotators.

approach in adversarial learning setups [40], [41].

### B. Real data

Further tests were conducted using real datasets. In this case, in addition to other ensemble learning algorithms, the proposed methods are also compared to the single best annotator, that is the classifier that exhibited the highest accuracy. For all experiments, a collection of $M=15$ classification algorithms from MATLAB's machine learning toolbox were trained, each on a different randomly selected subset of the dataset. Afterwards, the algorithms provided labels for all data in each dataset. The classification algorithms considered were $k$-nearest neighbor classifiers, for varying number of neighbors $k$ and different distance measures; support vector machine classifiers, utilizing different kernels; and decision trees with varying depth. The

| Algorithm | $N=100$ | $N=1000$ | $N=10^4$ | $N=10^5$ |
|---|---|---|---|---|
| KOS | 0.013 | 0.004 | 0.005 | 0.05 |
| EigenRatio | 0.003 | 0.002 | 0.005 | 0.03 |
| TE | 0.003 | 0.001 | 0.012 | 0.10 |
| SML | 0.04 | 0.09 | 0.76 | 11.98 |
| EM + MV | 0.01 | 0.02 | 0.12 | 1.47 |
| EM + Spectral | 1.48 | 1.55 | 1.58 | 3.00 |
| Alg. 2 | 1.82 | 2.32 | 2.05 | 3.01 |

TABLE III
CLASSIFICATION TIME (IN SECONDS) FOR A SYNTHETIC DATASET WITH $K=2$, PRIOR PROBABILITIES $\boldsymbol{\pi} = [0.5856, 0.4144]^\top$ AND $M=10$ ANNOTATORS.

real datasets considered are the MNIST dataset [42], and 5 UCI datasets [43]: the CoverType database, the PokerHand dataset, the Connect-4 dataset, the Magic dataset and the Dota 2 dataset. MNIST contains $N=70,000$ $28\times 28$ images of handwritten digits, each belonging to one of $K=10$ classes (one per digit). For this dataset, each classification algorithm was trained on subsets of $2,000$ instances. The CoverType dataset consists of $N=581,012$ data belonging to $K=7$ classes. Each cluster corresponds to a different forest cover type. Data are vectors of dimension $D=54$ that contain cartographic variables, such as soil type, elevation, hillshade etc. Here, each classification algorithm was trained on a subset of $1,000$ instances. The PokerHand database contains $N=10^6$ data belonging to $K=10$ classes. Each datum is a 5-card hand drawn from a deck of 52 cards, with each card being described by its rank and suit (spades, hearts,

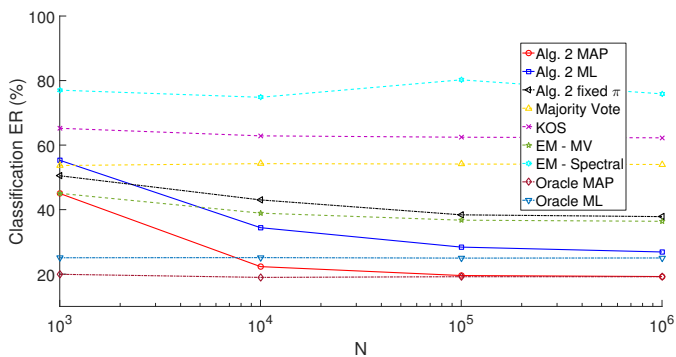

Fig. 4. Classification ER for a synthetic dataset with $K = 7$ classes, priors $\boldsymbol{\pi} = [0.2347, 0.0230, 0.0705, 0.1477, 0.2659, 0.0043, 0.2539]^\top$ and $M = 10$ annotators.

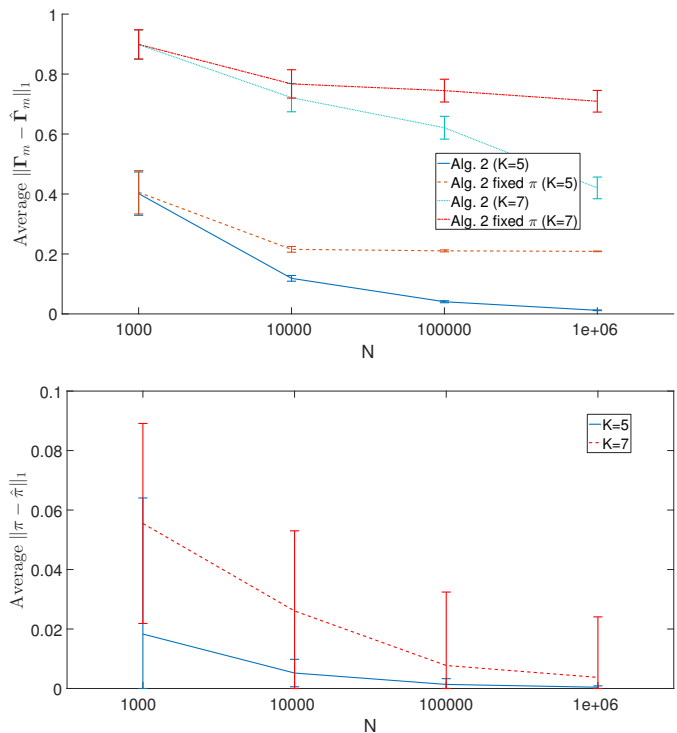

Fig. 5. Average estimation errors of confusion matrices (top); and prior probabilities (bottom) for two synthetic datasets with $K = 5$ and $K = 7$ classes and $M = 10$ annotators

diamonds, and clubs). Each class represents a valid Poker hand. For this experiment the 3 most prevalent classes are considered. Here, each classification algorithm was trained on a subset of $10,000$ instances. Connect-4 contains $N = 67,557$ vectors of size $42 \times 1$, each representing the possible positions in a connect-4 game. These vectors belong to one of $K = 3$ classes, indicating whether the first player is in a position to win, lose, or, tie the game. Here, each classification algorithm was trained on a subset of 300 instances. The Magic dataset contains $N = 19,020$ data captured by ground-based atmospheric Cherenkov gamma-ray detector. The dataset contains $K = 2$ classes, each indicating the presence or absence of Gamma rays. For this dataset, each classification algorithm was trained on subsets of 100 instances. The Dota 2 dataset contains $N = 102,944$ data, corresponding to different Dota

| Algorithm | $N = 1000$ | $N = 10^4$ | $N = 10^5$ | $N = 10^6$ |
|---|---|---|---|---|
| KOS | 0.016 | 0.02 | 0.17 | 2.03 |
| EM + MV | 0.04 | 0.27 | 3.43 | 37.27 |
| EM + Spectral | 119.35 | 124.94 | 119.35 | 160.54 |
| Alg. 2 | 28.27 | 40.23 | 36.08 | 47.17 |
| Alg. 2 fixed $\boldsymbol{\pi}$ | 13.34 | 6.23 | 6.11 | 18.16 |

TABLE IV
CLASSIFICATION TIME (IN SECONDS) FOR A SYNTHETIC DATASET WITH $K = 5$ CLASSES, PRIORS $\boldsymbol{\pi} = [0.2404, 0.2679, 0.0731, 0.1950, 0.2236]^\top$ AND $M = 10$ ANNOTATORS.

| Algorithm | $N = 1000$ | $N = 10^4$ | $N = 10^5$ | $N = 10^6$ |
|---|---|---|---|---|
| KOS | 0.017 | 0.025 | 0.23 | 2.83 |
| EM + MV | 0.05 | 0.30 | 4.80 | 48.87 |
| EM + Spectral | 619.61 | 616.47 | 621.30 | 676.95 |
| Alg. 2 | 46.19 | 52.66 | 54.50 | 69.99 |
| Alg. 2 fixed $\boldsymbol{\pi}$ | 34.94 | 38.88 | 39.11 | 40.17 |

TABLE V
CLASSIFICATION TIME (IN SECONDS) FOR A SYNTHETIC DATASET WITH $K = 7$ CLASSES, PRIORS $\boldsymbol{\pi} = [0.2347, 0.0230, 0.0705, 0.1477, 0.2659, 0.0043, 0.2539]^\top$ AND $M = 10$ ANNOTATORS.

2 games played, between two teams of 5 players. The dataset is split into $K = 2$ classes, corresponding to the team that won the game. Each datum consists of the starting parameters of each game, such as the game type (ranked or amateur) and which heroes were chosen from the players. Finally, for this dataset, each classification algorithm was trained on subsets of $5,000$ instances.

Table VIII lists classification ER results for the real data experiments. For most datasets, the proposed approaches outperform the competing alternatives, as well as the single-best classifier. For the MNIST dataset the EM methods of [22] outperform our approaches. Nevertheless, Alg. 1 comes very close to the performance of the EM schemes and if the confusion matrix estimates $\{\hat{\boldsymbol{\Gamma}}_m\}_{m=1}^M$ of Alg. 2 are refined using EM, we also reach a classification ER of $6.23\%$.

### C. Crowdsourcing data

In this section, the proposed scheme of Sec. III-F is evaluated on crowdsourcing data. The datasets considered are the Adult dataset [44], the TREC dataset [45] and the Bird

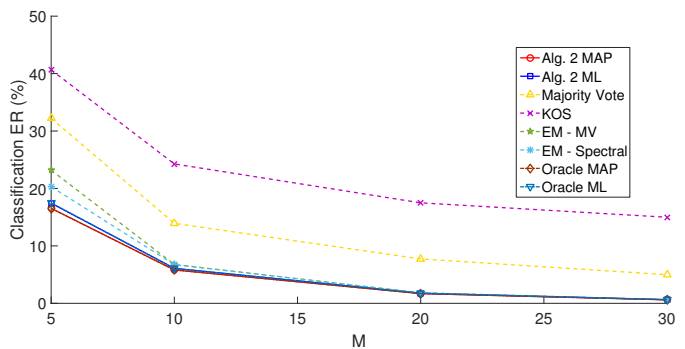

Fig. 6. Classification ER for a synthetic dataset with $K = 3$ classes, priors $\boldsymbol{\pi} = [0.2318, 0.4713, 0.2969]^\top$ and $N = 10^6$ data.

| Dataset | K | Single best | MV | EigenRatio | TE | SML | KOS | EM + MV | EM + Spectral | Alg. 2 MAP | Alg. 2 ML |
|---|---|---|---|---|---|---|---|---|---|---|---|
| MNIST | 10 | 7.29 | 7.0986 | - | - | - | 9.84 | **6.23** | **6.23** | 6.3986 | 6.3843 |
| CoverType | 7 | 29.89 | 28.642 | - | - | - | 31.13 | 58.68 | 95.62 | **28.574** | 28.913 |
| PokerHand | 3 | 41.95 | 43.365 | - | - | - | 49.62 | 53.62 | 78.38 | **39.436** | 39.339 |
| Connect-4 | 3 | 29.17 | 31.636 | - | - | - | 32.33 | 44.27 | 61.20 | **26.176** | 26.86 |
| Magic | 2 | 21.32 | 21.73 | 26.25 | 26.28 | 21.27 | 21.29 | 21.17 | 21.14 | **20.77** | 20.98 |
| Dota 2 | 2 | 41.27 | 42.174 | 45.55 | 45.75 | 40.568 | 40.59 | 40.80 | 59.19 | **40.497** | 40.549 |

TABLE VIII
CLASSIFICATION ER FOR REAL DATA EXPERIMENTS WITH $M = 15$.

| Algorithm | $M = 5$ | $M = 10$ | $M = 20$ | $M = 30$ |
|---|---|---|---|---|
| KOS | 0.44 | 0.96 | 4.13 | 5.29 |
| EM + MV | 11.48 | 21.67 | 41.88 | 62.19 |
| EM + Spectral | 21.92 | 32.77 | 53.88 | 75.24 |
| Alg. 2 | 4.85 | 15.43 | 83.73 | 271.71 |

TABLE VI
CLASSIFICATION TIME (IN SECONDS) FOR A SYNTHETIC DATASET WITH $K = 3$ CLASSES, PRIORS $\boldsymbol{\pi} = [0.2318, 0.4713, 0.2969]^\top$ AND $N = 10^6$ DATA.

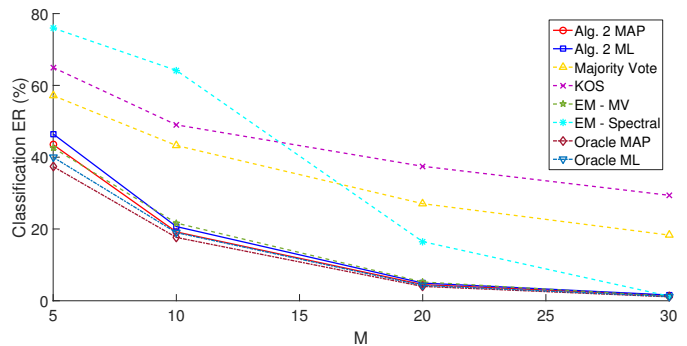

Fig. 8. Classification ER for a synthetic dataset with $K = 5$ classes, priors $\boldsymbol{\pi} = [0.3596, 0.1553, 0.1229, 0.3258, 0.0364]^\top$ and $N = 5,000$ data.

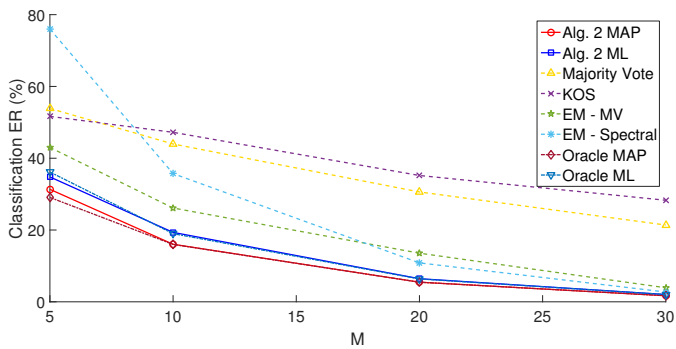

Fig. 7. Classification ER for a synthetic dataset with $K = 5$ classes, priors $\boldsymbol{\pi} = [0.3596, 0.1553, 0.1229, 0.3258, 0.0364]^\top$ and $N = 10^6$ data.

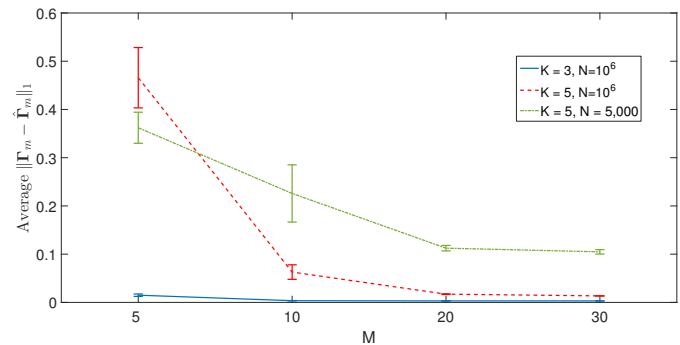

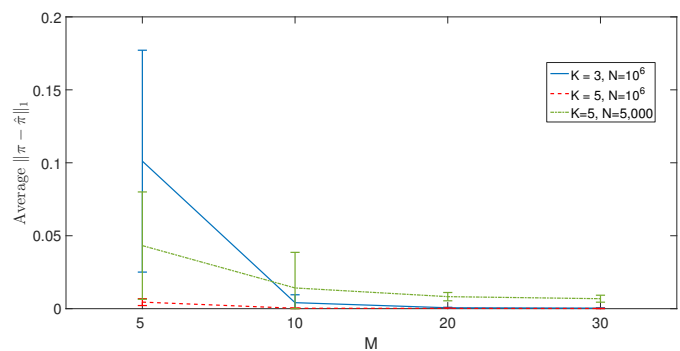

Fig. 9. Average estimation errors of confusion matrices (top); and prior probabilities (bottom) for two synthetic datasets with $K = 3$ and $K = 5$ classes and $N = 10^6$ data, and a synthetic dataset with $K = 5$ classes and $N = 5,000$ data.

dataset [46]. In most datasets, only a small set of ground-truth labels was available, and the performance of each method was evaluated on this set.

For the Adult dataset, annotators were tasked with classifying $N = 11,028$ websites into $K = 4$ different classes, using Amazon's Mechanical Turk [5]. The 4 classes correspond to different levels of adult content of a website. To maintain reasonable computational complexity, we only considered annotators that had given labels for all 4 classes and provided labels for more than 370 websites.

For the TREC dataset, annotators from Amazon's Mechanical Turk [5] were tasked with classifying $N = 19,033$ websites into $K = 2$ classes: "relevant" or "irrelevant" to some search queries. Again, to maintain reasonable computational complexity for our approach, we only considered annotators that had given labels for both classes and provided labels for more than 708 websites.

| Algorithm | $M = 5$ | $M = 10$ | $M = 20$ | $M = 30$ |
|---|---|---|---|---|
| KOS | 0.85 | 1.90 | 8.99 | 11.11 |
| EM + MV | 18.47 | 34.68 | 67.14 | 99.82 |
| EM + Spectral | 136.30 | 153.35 | 186.99 | 221.50 |
| Alg. 2 | 12.92 | 28.89 | 150.33 | 471.22 |

TABLE VII
CLASSIFICATION TIME (IN SECONDS) FOR A SYNTHETIC DATASET WITH $K = 5$ CLASSES, PRIORS $\boldsymbol{\pi} = [0.3596, 0.1553, 0.1229, 0.3258, 0.0364]^\top$ AND $N = 10^6$ DATA.

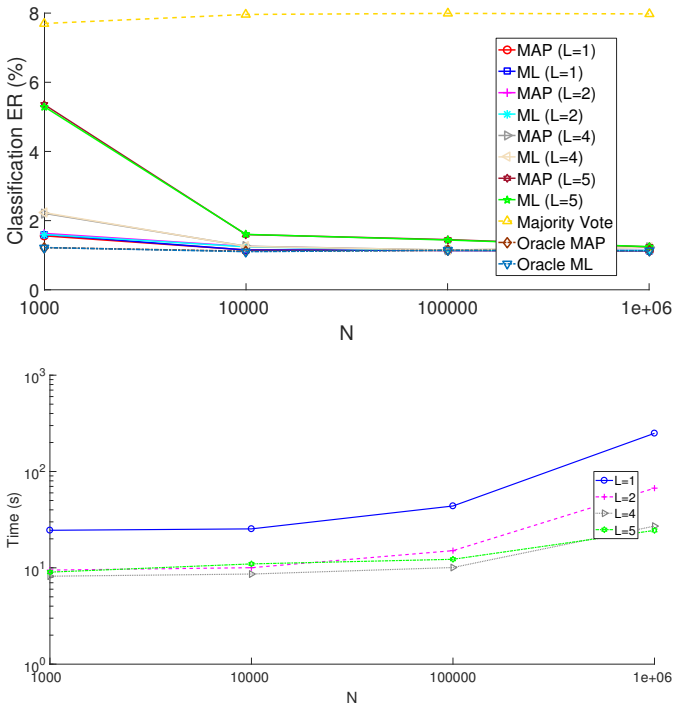

Fig. 10. Classification ER (top); and time (in seconds) (bottom) for a synthetic dataset with $K = 3$ classes, priors $\boldsymbol{\pi} = [0.3096, 0.3416, 0.3488]^\top$, $M = 30$ annotators for varying number of data $N$ and annotator groups $L$.

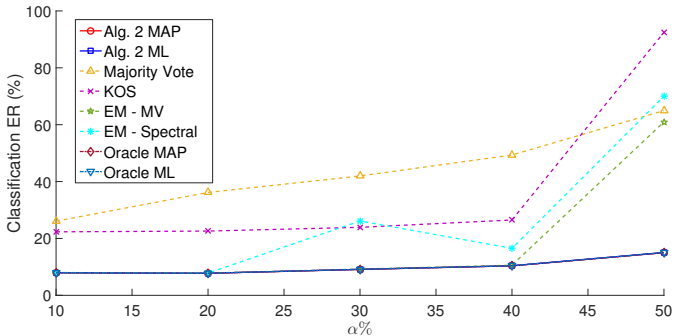

Fig. 11. Classification ER for a synthetic dataset with $K = 3$ classes, priors $\boldsymbol{\pi} = [0.31, 0.34, 0.35]^\top$, $N = 10^6$, $M = 10$ annotators and varying percentage of adversarial annotators $\alpha$.

For the bird dataset, annotators from Amazon's Mechanical Turk were tasked with classifying $N = 108$ images of birds into $K = 2$ classes: "Indigo Bunting" or "Blue Grosbeak".

Table IX lists classification ER for the two crowdsourcing experiments. The column "Labels" denotes the number of ground-truth labels available. As with the previous experiments, our approach exhibits lower classification ER than the competing alternatives, in both multiclass and binary classification settings.

## VI. CONCLUSIONS AND FUTURE DIRECTIONS

This paper introduced a novel approach to blind ensemble and crowdsourced classification that relies solely on annotator responses to assess their quality and combine their answers. Compact expressions of annotator moments, based on PARAFAC tensor decompositions were derived, and a novel moment matching scheme was developed using AO-ADMM. The performance of the novel algorithm was evaluated on real and synthetic data.

Several interesting research venues open up: i) Distributed and online implementations of the proposed algorithm to facilitate truly large-scale ensemble classification; ii) multiclass ensemble classification with dependent classifiers, along the lines of [47]; iii) ensemble clustering and regression; and iv) further investigation into the theoretical and practical implications of adversarial annotators along with possible remedies.

## APPENDIX A
## ALGORITHM DERIVATION

### A. ADMM subproblem for $\boldsymbol{\pi}$

Consider the following problem that is equivalent to (22)

$$\min_{\boldsymbol{\pi},\boldsymbol{\phi}} \; g_{N,\boldsymbol{\pi}}(\boldsymbol{\phi}) + \rho_{\mathcal{C}_p}(\boldsymbol{\pi}) \tag{28}$$
$$\text{s.to} \quad \boldsymbol{\pi} = \boldsymbol{\phi}$$

where $\boldsymbol{\phi}$ is an auxiliary variable used to capture the smooth part of the optimization problem, and $\rho_{\mathcal{C}_p}$ is an indicator function for the constraints of (22), namely

$$\rho_{\mathcal{C}_p}(\boldsymbol{u}) := \begin{cases} 0 & \text{if } \boldsymbol{u} \in \mathcal{C}_p \\ \infty & \text{otherwise.} \end{cases} \tag{29}$$

The augmented Lagrangian of (28) is then

$$\ell = g_{N,\boldsymbol{\pi}}(\boldsymbol{\phi}) + \rho_{\mathcal{C}_p}(\boldsymbol{\pi}) + \frac{\lambda}{2}\|\boldsymbol{\pi} - \boldsymbol{\phi} + \boldsymbol{\delta}\|_2^2 \tag{30}$$

where the $K \times 1$ vector $\boldsymbol{\delta}$ contains the scaled Lagrange multipliers for subproblem (22). Per ADMM iteration, (30) is minimized w.r.t. $\boldsymbol{\phi}$ and $\boldsymbol{\pi}$ before performing a gradient ascent step for $\boldsymbol{\delta}$. Specifically, the update for $\boldsymbol{\phi}$ at iteration $i+1$ is obtained by setting the gradient of $\ell$ w.r.t. $\boldsymbol{\phi}$ to $\mathbf{0}$, and solving for $\boldsymbol{\phi}$; that is,

$$\begin{aligned}
&\bigg((\lambda+\nu)\mathbf{I} + \sum_{m=1}^{M} \boldsymbol{\Gamma}_m^\top \boldsymbol{\Gamma}_m + \sum_{\substack{m=1 \\ m'>m}}^{M} \mathbf{K}_{m'm}^\top \mathbf{K}_{m'm} \\
&\quad + \sum_{\substack{m=1 \\ m'>m \\ m''>m'}}^{M} (\boldsymbol{\Gamma}_{m''} \odot \mathbf{K}_{m'm})^\top (\boldsymbol{\Gamma}_{m''} \odot \mathbf{K}_{m'm})\bigg) \boldsymbol{\phi}[i+1] \\
&= \sum_{m=1}^{M} \boldsymbol{\Gamma}_m^\top \boldsymbol{\mu}_m + \sum_{\substack{m=1 \\ m'>m}}^{M} \mathbf{K}_{m'm}^\top \boldsymbol{s}_{mm'} + \nu\boldsymbol{\pi}^{(\text{prev})} \\
&\quad + \lambda(\boldsymbol{\pi}[i] + \boldsymbol{\delta}[i]) + \sum_{\substack{m=1 \\ m'>m \\ m''>m'}} (\boldsymbol{\Gamma}_{m''} \odot \mathbf{K}_{m'm})^\top \boldsymbol{t}_{mm'm''},
\end{aligned} \tag{31}$$

where $\mathbf{K}_{mm'} := \boldsymbol{\Gamma}_m \odot \boldsymbol{\Gamma}_{m'}$. Brackets here indicate ADMM iteration indices. Accordingly, the update for $\boldsymbol{\pi}$ is given by

$$\boldsymbol{\pi}[i+1] = P_{\mathcal{C}_p}\big(\boldsymbol{\phi}[i+1] - \boldsymbol{\delta}[i]\big) \tag{32}$$

where $P_{\mathcal{C}_p}$ is the projection operator onto the convex set $\mathcal{C}_p$; that is, $\boldsymbol{\phi}[i+1] - \boldsymbol{\delta}[i]$ is projected onto the probability simplex.

| Dataset | N | K | M | Labels | MV | EigenRatio | TE | SML | KOS | EM + MV | EM + Spectral | Alg. 2 MAP | Alg. 2 ML |
|---|---|---|---|---|---|---|---|---|---|---|---|---|---|
| Adult | 11,028 | 4 | 38 | 347 | 36.023 | - | - | - | 80.98 | 40.63 | 38.90 | **33.429** | **34.87** |
| TREC | 19,033 | 2 | 23 | 2,275 | 50.002 | 43.34 | 48.97 | 48.44 | 54.68 | 56.04 | 40.62 | **37.846** | **39.824** |
| Bird | 108 | 2 | 39 | 108 | 24.07 | 27.78 | 17.59 | 11.11 | 11.11 | 11.11 | **10.19** | **10.19** | **10.19** |

TABLE IX
CLASSIFICATION ER FOR CROWDSOURCING DATA EXPERIMENTS.

This projection can be performed using efficient methods [48]. Finally, a gradient ascent step is performed for $\boldsymbol{\delta}$ as

$$\boldsymbol{\delta}[i+1] = \boldsymbol{\delta}[i] + \boldsymbol{\pi}[i+1] - \boldsymbol{\phi}[i+1]. \quad (33)$$

Note that products of the form $\mathbf{K}_{m'm}^\top \mathbf{K}_{m'm} = (\boldsymbol{\Gamma}_m \odot \boldsymbol{\Gamma}_{m'})^\top (\boldsymbol{\Gamma}_m \odot \boldsymbol{\Gamma}_{m'})$ can be efficiently computed by using the following observation: $(\boldsymbol{\Gamma}_m \odot \boldsymbol{\Gamma}_{m'})^\top (\boldsymbol{\Gamma}_m \odot \boldsymbol{\Gamma}_{m'}) = (\boldsymbol{\Gamma}_m^\top \boldsymbol{\Gamma}_m) * (\boldsymbol{\Gamma}_{m'}^\top \boldsymbol{\Gamma}_{m'})$, where $*$ denotes the elementwise matrix product [11]. In addition, the products $\boldsymbol{\Gamma}_m^\top \boldsymbol{\Gamma}_m$ do not have to be explicitly computed each time (28) is solved, as they can be cached every time (34) is solved. As suggested in [28], the maximum number of ADMM iterations, $I$, for each subproblem can be set to be small, e.g. $I = 10$.

### B. ADMM subproblem for $\boldsymbol{\Gamma}_m$

Proceeding along similar lines with the previous subsection, consider the following problem which is equivalent to (23)

$$\min_{\boldsymbol{\Gamma}_m, \boldsymbol{\Phi}} \quad \bar{g}_{N,m}(\boldsymbol{\Gamma}_m, \boldsymbol{\Phi}) \quad (34)$$
$$\text{s.to} \quad \boldsymbol{\Gamma}_m = \boldsymbol{\Phi}^\top$$

where $\boldsymbol{\Phi}$ is an auxiliary variable used to capture the smooth part of the optimization problem in (23), and

$$\bar{g}_{N,m}(\boldsymbol{\Gamma}_m, \boldsymbol{\Phi}) = g_{N,m}(\boldsymbol{\Phi}^\top) + \rho_{\mathcal{C}}(\boldsymbol{\Gamma}_m).$$

The augmented Lagrangian of (34) is then

$$\ell' = \bar{g}_{N,m}(\boldsymbol{\Gamma}_m, \boldsymbol{\Phi}) + \frac{\lambda}{2} \|\boldsymbol{\Gamma}_m - \boldsymbol{\Phi}^\top + \boldsymbol{\Delta}_m\|_F^2 \quad (35)$$

where the $K \times K$ matrix $\boldsymbol{\Delta}_m$ contains the scaled Lagrange multipliers for subproblem (23), and $\lambda$ is a positive scalar. As in the previous section, per ADMM iteration, (35) is minimized with respect to (w.r.t.) $\boldsymbol{\Phi}$ and $\boldsymbol{\Gamma}_m$ before performing a gradient ascent step for $\boldsymbol{\Delta}_m$. Specifically, the update for $\boldsymbol{\Phi}$ at iteration $i+1$ is obtained by setting the gradient of $\ell'$ w.r.t. $\boldsymbol{\Phi}$ to $\mathbf{0}$, and solving for $\boldsymbol{\Phi}$. Since $\mathbf{S}_{m'm} = \mathbf{S}_{mm'}^\top$ and $\boldsymbol{\Pi} = \boldsymbol{\Pi}^\top$, it is easy to see that the update w.r.t. $\boldsymbol{\Phi}$ can be expressed as

$$\left( (\lambda + \nu)\mathbf{I} + \boldsymbol{\pi}\boldsymbol{\pi}^\top + \sum_{m' \neq m}^M \boldsymbol{\Pi}\boldsymbol{\Gamma}_{m'}^\top \boldsymbol{\Gamma}_{m'}\boldsymbol{\Pi} \right.$$
$$\left. + \sum_{\substack{m'>m \\ m''>m'}} \boldsymbol{\Pi}\mathbf{K}_{m''m'}^\top \mathbf{K}_{m''m'}\boldsymbol{\Pi} \right) \boldsymbol{\Phi}[i+1]$$
$$= \boldsymbol{\pi}\boldsymbol{\mu}_m^\top + \sum_{m' \neq m}^M \boldsymbol{\Pi}\boldsymbol{\Gamma}_{m'}^\top \mathbf{S}_{m'm} + \sum_{\substack{m'>m \\ m''>m'}} \boldsymbol{\Pi}\mathbf{K}_{m''m'}^\top \mathbf{T}_{mm'm''}^{(1)}$$
$$+ \nu \boldsymbol{\Gamma}_m^{(\text{prev})\top} + \lambda(\boldsymbol{\Gamma}_m[i] + \boldsymbol{\Delta}_m[i])^\top. \quad (36)$$

Accordingly, the update for $\boldsymbol{\Gamma}_m$ is given by

$$\boldsymbol{\Gamma}_m[i+1] = P_{\mathcal{C}}\left(\boldsymbol{\Phi}^\top[i+1] - \boldsymbol{\Delta}_m[i]\right) \quad (37)$$

where $P_{\mathcal{C}}$ is the projection operator onto the convex set $\mathcal{C}$ with each column of $\boldsymbol{\Phi}^\top[i+1] - \boldsymbol{\Delta}_m[i]$ projected onto the probability simplex. Finally, a gradient ascent step is performed per $\boldsymbol{\Delta}_m$, as follows

$$\boldsymbol{\Delta}_m[i+1] = \boldsymbol{\Delta}_m[i] + \boldsymbol{\Gamma}_m[i+1] - \boldsymbol{\Phi}^\top[i+1]. \quad (38)$$

### C. Algorithm complexity

For the ADMM subproblems of Apps. A-A and A-B the complexity per iteration is dominated by the matrix inversions required in (31) and (36) respectively, that is $\mathcal{O}(K^3)$. However, in order to instantiate the left- and right-hand sides of (31), $\mathcal{O}(M^3K^2)$ and $\mathcal{O}(M^3K^4)$ operations are required respectively. These operations have to be performed only once and cached to be used in each iteration. The increased complexity of the right-hand side is due to the matricized tensor times Khatri-Rao product (MTTKRP) $(\boldsymbol{\Gamma}_{m''} \odot \mathbf{K}_{m'm})^\top \boldsymbol{t}_{mm'm''}$. These MTTKRPs however, can be computed efficiently due to the Khatri-Rao structure, and are easily parallelizable, see e.g. [49]. This brings the overall complexity of App. A-A to $\mathcal{O}(M^3K^4 + IK^3)$, with $I$ denoting the number of ADMM iterations. Accordingly, the operations required to instantiate the left- and right-hand sides of (36) are $\mathcal{O}(M^2K^2)$ and $\mathcal{O}(M^2K^4)$ respectively. This brings the total complexity of App. A-B to $\mathcal{O}(M^2K^4 + IK^3)$. As the number of iterations for the ADMM algorithms of Apps. A-A and A-B is set to be small the overall computational complexity of Alg. 1 is $\mathcal{O}(I_T M^3 K^4)$, where $I_T$ is the number of AO-ADMM iterations required until convergence.

Furthermore, the number of tensors $\underline{T}_{mm'm''}$ required to solve (21) is $\binom{M}{3}$, while the number of matrices $\mathbf{S}_{mm'}$ required is $\binom{M}{2}$, and the number of vectors $\boldsymbol{\mu}_m$ is $M$. Thus, for $K$ classes, the memory needed for storing all the tensors, matrices and vectors involved is $\mathcal{O}\left(\binom{M}{3}K^3 + \binom{M}{2}K^2 + MK\right)$. Finally, computing the cross-correlation tensors, matrices and mean vectors of annotators incurs a complexity of $\mathcal{O}(M^3KN)$ as each of the annotator response matrices $\{\mathbf{F}_m\}_{m=1}^M$ is of size $K \times N$ and has $N$ nonzero entries.

## APPENDIX B
## PROOFS

**Proof of Lemma 1.** Suppose that $\text{rank}(\boldsymbol{\Gamma}_m) = \text{rank}(\boldsymbol{\Gamma}_{m'}) = \text{rank}(\boldsymbol{\Gamma}_{m''}) = K$, for some $m \neq m', m''$ and $m' \neq m''$. Then by [11, Thm. 2] the decomposition of $\underline{\Psi}_{mm'm''}$ is *essentially unique*. Invoking [36, Prop 4.10] the joint tensor

decomposition of (21) is *essentially unique*, meaning the solutions of (21) will be of the form

$$\hat{\mathbf{\Gamma}}_m = \mathbf{\Gamma}_m^* \mathbf{P} \mathbf{\Lambda}_m, \quad m = 1, \ldots, M, \quad \hat{\boldsymbol{\pi}} = \mathbf{\Lambda} \mathbf{P}^\top \boldsymbol{\pi}^*$$

where $\mathbf{P}$ is a permutation matrix, and $\{\mathbf{\Lambda}_m\}_{m=1}^M$, $\mathbf{\Lambda}$ are diagonal scaling matrices such that $\mathbf{\Lambda}_m \mathbf{\Lambda}_{m'} \mathbf{\Lambda}_{m''} = \mathbf{\Lambda}^{-1}$, for $m \neq m', m'', m' \neq m''$. Since $\{\hat{\mathbf{\Gamma}}_m\}$ and $\hat{\boldsymbol{\pi}}$ are the solutions to (21), they must satisfy the constraints of the optimization problem; that is $\hat{\mathbf{\Gamma}}_m \in \mathcal{C}$ $m = 1, \ldots, M$ and $\hat{\boldsymbol{\pi}} \in \mathcal{C}_p$.

Since $\mathbf{\Gamma}_m^{*\top} \mathbf{1} = \mathbf{1}$ for all $m$, and $\mathbf{P}^\top \mathbf{1} = \mathbf{1}$, we have

$$\hat{\mathbf{\Gamma}}_m^\top \mathbf{1} = \mathbf{1} \Rightarrow \mathbf{\Lambda}_m \mathbf{P}^\top \mathbf{\Gamma}_m^{*\top} \mathbf{1} = \mathbf{1} \Rightarrow \mathbf{\Lambda}_m \mathbf{1} = \mathbf{1} \ m = 1, \ldots, M$$

which implies that $\mathbf{\Lambda}_m = \mathbf{I}$ for $m = 1, \ldots, M$. Since $\mathbf{\Lambda}_m \mathbf{\Lambda}_{m'} \mathbf{\Lambda}_{m''} = \mathbf{\Lambda}^{-1}$, for $m \neq m', m'', m' \neq m''$, we arrive at $\mathbf{\Lambda} = \mathbf{I}$. Thus, the constraints of (21) solve the possible scaling ambiguities. Letting $\hat{\mathbf{P}} = \mathbf{P}^\top = \mathbf{P}^{-1}$, we arrive at the statement of the lemma. □

**Proof of Theorem 1.** For notational convenience, collect all optimization variables in $\boldsymbol{\theta}$, and denote the aggregated constraint set as $\bar{\mathcal{C}}$. Note that $\bar{\mathcal{C}}$ is a compact set, since the probability simplex is compact and $\bar{\mathcal{C}}$ is an intersection of simplexes. Since $h_N(\boldsymbol{\theta})$ is continuous and $\bar{\mathcal{C}}$ is compact, $h_N(\boldsymbol{\theta})$ is uniformly continuous on $\bar{\mathcal{C}}$, that is, $\forall \varepsilon > 0$ there exists a neighborhood $\mathcal{V}$ of $\tilde{\boldsymbol{\theta}}$ such that

$$\sup_{\boldsymbol{\theta} \in \mathcal{V} \cap \bar{\mathcal{C}}} |h_N(\boldsymbol{\theta}) - h_N(\tilde{\boldsymbol{\theta}})| < \varepsilon/2. \quad (39)$$

Due to the compactness of $\bar{\mathcal{C}}$ there exist a finite number of points $\boldsymbol{\theta}_1, \ldots, \boldsymbol{\theta}_L \in \bar{\mathcal{C}}$, with corresponding neighborhoods $\mathcal{V}_1, \ldots, \mathcal{V}_L$ that cover $\bar{\mathcal{C}}$, that is

$$\sup_{\boldsymbol{\theta} \in \mathcal{V}_\ell \cap \bar{\mathcal{C}}} |h_N(\boldsymbol{\theta}) - h_N(\boldsymbol{\theta}_\ell)| < \varepsilon/2, \quad \text{for } \ell = 1, \ldots, L. \quad (40)$$

Invoking the LLN, it is straightforward to show that, for sufficiently large $N$, w.p. 1

$$|h_N(\boldsymbol{\theta}_\ell) - h_\infty(\boldsymbol{\theta}_\ell)| < \varepsilon/2, \quad \text{for } \ell = 1, \ldots, L. \quad (41)$$

Using the triangle inequality along with (40), and (41) we have

$$\sup_{\boldsymbol{\theta} \in \bar{\mathcal{C}}} |h_N(\boldsymbol{\theta}) - h_\infty(\boldsymbol{\theta})| < \varepsilon, \quad (42)$$

that is, for sufficiently large $N$, $h_N$ converges uniformly to $h_\infty$ on $\bar{\mathcal{C}}$. Then, by [38, Thm. 5.3] we have that $D(\mathcal{S}_N, \mathcal{S}_*) \to 0$ as $N \to \infty$. □

**Proof of Theorem 2.** Let $\bar{L}(x|k) = L(x|k)\pi_k$, with $L(x|k)$ as defined in Sec. III-A. Then the average probability of error of the MAP detector can be expressed as

$$\mathrm{P_e} = \sum_{k=1}^K \mathrm{P_{e,k}} \pi_k \quad (43)$$

where $\mathrm{P_{e,k}} = \Pr(\bar{L}(x|k) < \bar{L}(x|k'), k' \neq k | Y = k)$. By applying a union bound on $\mathrm{P_{e,k}}$ it is easy to show that

$$\mathrm{P_{e,k}} \leq \sum_{k' \neq k} \Pr(\bar{L}(x|k) < \bar{L}(x|k') | Y = k). \quad (44)$$

Defining $\mathrm{P}_{\bar{L}}(k, k') := \Pr(\bar{L}(x|k) < \bar{L}(x|k') | Y = k)$, substituting (44) in (43) and grouping terms we have

$$\mathrm{P_e} \leq \sum_{k=1}^K \sum_{k' > k} \pi_k \mathrm{P}_{\bar{L}}(k, k') + \pi_{k'} \mathrm{P}_{\bar{L}}(k', k). \quad (45)$$

Consider now the binary hypothesis testing problem between classes $k$ and $k' \neq k$. The average probability of error of a MAP detector for the binary problem is

$$\mathrm{P_e}(k, k') = \frac{\pi_k}{\pi_k + \pi_{k'}} \mathrm{P}_{\bar{L}}(k, k') + \frac{\pi_{k'}}{\pi_k + \pi_{k'}} \mathrm{P}_{\bar{L}}(k', k). \quad (46)$$

Then

$$\pi_k \mathrm{P}_{\bar{L}}(k, k') + \pi_{k'} \mathrm{P}_{\bar{L}}(k', k)$$
$$= (\pi_k + \pi_{k'}) \mathrm{P_e}(k, k') \leq \mathrm{P_e}(k, k') \quad (47)$$

where the inequality is due to $\pi_k + \pi_{k'} \leq 1$. Combining (47) with (45) yields

$$\mathrm{P_e} \leq \sum_{k=1}^K \sum_{k' > k} \mathrm{P_e}(k, k'). \quad (48)$$

Therefore, we have upper bounded the average probability of error of our $M$-class hypothesis testing problem by the average error probabilities of binary hypothesis testing problems. For the binary hypothesis testing problem between classes $k$ and $k' \neq k$, collect all annotator responses in an $M \times 1$ vector $\tilde{\boldsymbol{f}}$ and define two complementary regions $\mathcal{R}$ and $\mathcal{R}^C$ as

$$\mathcal{R} = \{\tilde{\boldsymbol{f}} : \bar{L}(x|k) < \bar{L}(x|k')\} \quad (49a)$$
$$\mathcal{R}^C = \{\tilde{\boldsymbol{f}} : \bar{L}(x|k') < \bar{L}(x|k)\}. \quad (49b)$$

Upon defining $\tilde{\pi}_{k,k'} = \frac{\pi_k}{\pi_k + \pi_{k'}}$ and using (49), (46) can be rewritten as

$$\mathrm{P_e}(k, k') = \Pr(\tilde{\boldsymbol{f}} \in \mathcal{R} | Y = k) \tilde{\pi}_{k,k'} + \Pr(\tilde{\boldsymbol{f}} \in \mathcal{R}^C | Y = k') \tilde{\pi}_{k',k}$$
$$= \prod_{m=1}^M \Pr([\tilde{\boldsymbol{f}}]_m \in \mathcal{R}_m | Y = k) \tilde{\pi}_{k,k'}$$
$$+ \prod_{m=1}^M \Pr([\tilde{\boldsymbol{f}}]_m \in \mathcal{R}_m^C | Y = k') \tilde{\pi}_{k',k} \quad (50)$$

where the second equality follows from As. 1 and $\mathcal{R}_m, \mathcal{R}_m^C$ denote the subsets of $\mathcal{R}, \mathcal{R}^C$ corresponding to the $m$-th entry of $\tilde{\boldsymbol{f}}$, respectively. Now let

$$m^* = \arg\max_m \Pr([\tilde{\boldsymbol{f}}]_m \in \mathcal{R}_m | Y = k)^M \tilde{\pi}_{k,k'} \quad (51)$$
$$+ \Pr([\tilde{\boldsymbol{f}}]_m \in \mathcal{R}_m^C | Y = k')^M \tilde{\pi}_{k',k}$$

and define

$$\bar{\mathrm{P}}_e(k, k') = \Pr([\tilde{\boldsymbol{f}}]_{m^*} \in \mathcal{R}_{m^*} | Y = k)^M \tilde{\pi}_{k,k'}$$
$$+ \Pr([\tilde{\boldsymbol{f}}]_{m^*} \in \mathcal{R}_{m^*}^C | Y = k')^M \tilde{\pi}_{k',k}. \quad (52)$$

Clearly $\mathrm{P_e}(k, k') \leq \bar{\mathrm{P}}_e(k, k')$. From standard results in detection theory (52) can be bounded as [50], [51]

$$\bar{\mathrm{P}}_e(k, k') \leq \exp(-Md(p\|q)) \quad (53)$$

where $p := \Pr([\tilde{\boldsymbol{f}}]_{m^*} \in \mathcal{R}_{m^*} | Y = k)$, $q := \Pr([\tilde{\boldsymbol{f}}]_{m^*} \in \mathcal{R}_{m^*}^C | Y = k')$, and $d(p\|q)$ denotes the Chernoff information between pdfs $p$ and $q$. Combining (53) with (48) yields the claim of the theorem. □